\documentclass{article}
\usepackage{iclr2026_conference,times}
\usepackage[hyphens]{url}
\usepackage{graphicx}
\usepackage{caption}
\usepackage{amssymb}
\usepackage{amsmath}
\usepackage{multirow}
\usepackage{pifont}
\usepackage{booktabs}
\usepackage{makecell}
\usepackage{tikz}
\usepackage{pgfplots}
\pgfplotsset{compat=1.18}
\usepackage[hidelinks]{hyperref}
\usepackage{colortbl}
\definecolor{oursRow}{RGB}{252,238,240}

\definecolor{lotaRed}{RGB}{190,35,45}
\definecolor{methodBlue}{RGB}{58,92,160}
\definecolor{softGray}{RGB}{90,90,90}

\DeclareMathOperator{\Tr}{Tr}
\newcommand{\safeincludegraphics}[2][]{%
  \IfFileExists{#2}{\includegraphics[#1]{#2}}{%
    \fbox{\parbox[c][1.15in][c]{0.9\linewidth}{\centering Missing figure: \texttt{\detokenize{#2}}}}%
  }%
}

\title{LoTA-N2N: Local Trace Adaptation for Zero-Shot Self-Supervised Image Denoising}
\author{
Jintong Hu\textsuperscript{1} \quad
Bin Xia\textsuperscript{2} \quad
Junlin Liu\textsuperscript{3} \quad
Jiayue Liu\textsuperscript{4} \quad
Wenming Yang\textsuperscript{1,*}\\[4pt]
\textsuperscript{1}Tsinghua University,
\textsuperscript{2}The Chinese University of Hong Kong\\
\textsuperscript{3}University of Southern California,
\textsuperscript{4}Peking University\\[2pt]
\textsuperscript{*}Corresponding author
}
\date{}

\iclrfinalcopy

\begin{document}
\maketitle
\fancypagestyle{arxiv}{%
    \fancyhf{}%
    \fancyfoot[C]{\thepage}%
    \renewcommand{\headrulewidth}{0pt}%
    \renewcommand{\footrulewidth}{0pt}%
}
\pagestyle{arxiv}
\thispagestyle{arxiv}

\begin{abstract}
Single-image self-supervised denoising replaces unavailable clean targets with surrogate targets constructed from noisy observations. Its effectiveness therefore depends on how closely the surrogate objective remains aligned with supervised denoising, especially when noise is correlated, spatially nonstationary, or unknown. We express the discrepancy between a broad class of MSE-based self-supervised objectives and supervised MSE as a parameter-independent constant and a trace interaction between the surrogate-target residual and the prediction error. The corresponding gradient discrepancy is determined by the gradient of this interaction. This formulation provides a common view of paired-noise, blind-spot, weak-noise, re-corruption, and sub-image methods, while revealing that a small global interaction may conceal substantial positive and negative regional interactions through spatial cancellation. Building on these observations, we propose LoTA-N2N, a two-stage zero-shot adaptation framework. Stage~1 trains a denoiser on complementary sub-image pairs and freezes it to construct detached clean-sub-image proxies. Stage~2 estimates the residual--prediction interaction using these proxies and suppresses its patch-wise absolute magnitude. We show that the local construction prevents spatial cancellation and upper-bounds the magnitude of the corresponding global interaction. We further derive a population supervised-risk bound whose slack depends on the Stage~1 clean-estimation error. Experiments across natural, confocal, and X-ray images, complemented by iteration-matched controls, controlled noise shifts, and gradient diagnostics, show consistent gains over MSE-only adaptation under IID, spatially varying, and mixed noise. Overall, LoTA-N2N demonstrates that estimated local interaction and spatial cancellation control provide effective design principles for single-image self-supervised denoising without paired clean targets, repeated acquisitions, or a predefined re-corruption model.
\end{abstract}

\section{Introduction}
\label{sec1}

Deep image denoisers are typically trained against clean targets. In many imaging systems, however, aligned noisy-clean pairs are expensive, difficult to acquire, or unavailable. This is particularly common in microscopy, medical imaging, remote sensing, and real camera pipelines, where repeated acquisition may introduce motion, photobleaching, exposure variation, or other misalignment \cite{Zhang_2017,Zhang_2018,anwar2020real,guo2019convolutional}. Self-supervised denoising addresses this limitation by constructing supervision directly from noisy observations.

Existing methods differ mainly in how they construct surrogate supervision. Noise2Noise uses independently corrupted observations of the same scene \cite{lehtinen2018noise2noise}; blind-spot methods predict held-out pixels from spatial context \cite{krull2019noise2void,batson2019noise2self,laine2019highquality,lee2022apbsn}; sub-image methods construct proxy pairs from neighboring pixels \cite{huang2021neighbor2neighbor,mansour2023zeroshot}; and re-corruption methods synthesize additional noisy observations \cite{Xu_2020,9577798,zhang2022idr,Zou_2023_ICCV}. Noise2Same instead derives a self-supervised upper bound on supervised risk using a reconstruction term and a prediction-invariance discrepancy \cite{xie2020noise2same}. Despite their different constructions, these methods all require an observable self-supervised objective to remain sufficiently aligned with clean-image denoising. This alignment can deteriorate under correlated or spatially nonstationary noise, neighboring-pixel mismatch, an inaccurate re-corruption model, or an unreliable clean-image proxy.

To characterize this mismatch, we write the surrogate target as a latent clean target plus residual corruption. The resulting self-supervised MSE decomposes exactly into supervised MSE, a parameter-independent constant, and an interaction between the surrogate residual and the prediction error. Its gradient discrepancy from supervised learning is exactly twice the gradient of this interaction. This decomposition provides a common criterion under which independent noisy targets, blind-spot masking, weak-noise training, model-based re-corruption, and sub-image pairing can be understood as different sufficient mechanisms for suppressing the same interaction.

The use of a supervised-risk bound is not itself unique to our work. Noise2Same controls a masking-based prediction-invariance discrepancy during dataset-level self-supervised training. LoTA-N2N addresses a different setting and observable quantity: it estimates the residual--prediction interaction from a frozen teacher trained on the same noisy image and localizes this interaction during per-image sub-image adaptation. The central contribution is therefore not the MSE expansion alone, but the combination of estimated interaction, spatial cancellation control, and bound-motivated zero-shot adaptation.

A global interaction value is insufficient for this purpose because positive and negative regional interactions may cancel even when their local magnitudes remain large. Moreover, a small interaction at one parameter point does not imply a small gradient discrepancy. We therefore minimize patch-wise absolute interaction estimates to prevent spatial cancellation, while measuring first-order discrepancy separately as a diagnostic. The inaccessible true local interaction yields a population supervised-risk bound with an explicit estimation-error term. The scheduled practical loss is motivated by this bound but is not claimed to be the bound itself.

\begin{figure*}[t]
    \centering
    \begin{minipage}[t]{0.44\textwidth}
\centering
{\small\bfseries (a) Trace correction mechanism\par}
\vspace{0.6mm}
\resizebox{\linewidth}{!}{%
\begin{tikzpicture}[x=1cm,y=1cm]
    \path[use as bounding box] (0.00,0.00) rectangle (7.20,4.85);

    \node[
        draw=black!45,
        fill=black!2,
        rounded corners=2pt,
        line width=0.45pt,
        align=center,
        text width=6.45cm,
        minimum height=0.56cm,
        font=\small
    ] at (3.60,4.48)
        {$\mathcal{L}_{\mathrm{self\text{-}supervised}}
        = \mathcal{L}_{\mathrm{supervised}} + C - 2\mathcal{T}(\theta)$};

    \node[
        font=\scriptsize,
        text=softGray,
        align=center
    ] at (3.60,3.92)
        {Self-supervised risk approaches supervised risk when the trace term is small};

    \node[
        draw=methodBlue!75!black,
        fill=methodBlue!8,
        rounded corners=2pt,
        line width=0.45pt,
        align=center,
        text width=1.68cm,
        minimum height=0.78cm,
        font=\small
    ] (ss) at (0.98,3.00)
        {Self-supervised\\Loss};

    \node[
        draw=lotaRed!80!black,
        fill=lotaRed!8,
        rounded corners=2pt,
        line width=0.55pt,
        align=center,
        text width=1.58cm,
        minimum height=0.68cm,
        font=\small
    ] (gap) at (3.60,3.00)
        {Trace gap\\ $2\mathcal{T}(\theta)$};

    \node[
        draw=black!55,
        fill=black!3,
        rounded corners=2pt,
        line width=0.45pt,
        align=center,
        text width=1.68cm,
        minimum height=0.78cm,
        font=\small
    ] (sup) at (6.22,3.00)
        {Supervised\\Loss};

    \draw[->, line width=0.55pt, draw=black!65]
        (ss.east) -- (gap.west);

    \draw[<-, line width=0.55pt, draw=black!65]
        (gap.east) -- (sup.west);

    \node[
        draw=lotaRed!80!black,
        fill=lotaRed!7,
        rounded corners=2pt,
        line width=0.55pt,
        align=center,
        text width=5.95cm,
        minimum height=1.10cm,
        font=\small
    ] (lota) at (3.60,1.28)
        {LoTA-N2N suppresses local trace magnitude\\[1.6mm]
        $\displaystyle
        \frac{1}{M}\sum_m
        \left|\widehat{\mathcal{T}}^{(m)}(\theta)\right|$};

    \draw[->, line width=0.60pt, draw=lotaRed!75!black]
        (lota.north) -- node[
            right,
            midway,
            font=\scriptsize,
            text=lotaRed!85!black,
            xshift=1.0mm
        ] {reduces} (gap.south);

    \node[
        font=\scriptsize,
        text=softGray,
        align=center
    ] at (3.60,0.18)
        {No paired clean data, repeated acquisitions, or explicit noise model};

\end{tikzpicture}
}
\end{minipage}
    \hfill
    \begin{minipage}[t]{0.53\textwidth}
    \centering
    {\small\bfseries (b) Per-image gains over iteration-matched MSE\par}
    \vspace{0.2mm}
    \begin{tikzpicture}
    \begin{axis}[
        width=\linewidth,
        height=0.5\linewidth,
        ymin=0.0,
        ymax=1.25,
        xmin=0.45,
        xmax=3.55,
        ylabel={$\Delta$PSNR over MSE (dB)},
        xlabel={Noise setting},
        xtick={1,2,3},
        xticklabels={
            \makecell[c]{Gaussian\\[-0.2mm]$\sigma=20$},
            \makecell[c]{Poisson\\[-0.2mm]peak $=50$},
            \makecell[c]{Mixed noise\\[-0.2mm]peak $=50$\\[-0.2mm]$\sigma=20$}
        },
        tick label style={font=\small},
        xticklabel style={font=\footnotesize, align=center},
        label style={font=\small},
        xlabel style={yshift=-0.8mm},
        ymajorgrids=true,
        grid style={line width=.2pt, draw=gray!25},
        axis x line*=bottom,
        axis y line*=left,
        axis line style={draw=black!70},
        tick style={draw=black!70},
        clip=false
    ]

    \addplot[dashed, draw=black!55, line width=0.45pt]
        coordinates {(0.5,0) (3.5,0)};

    \addplot[
        only marks, mark=*, mark size=1.8pt,
        draw=lotaRed!80!black, fill=lotaRed!65, opacity=0.85
    ] coordinates {
        (0.82,0.348) (0.88,0.529) (0.94,0.392) (1.00,0.333)
        (1.06,0.889) (1.12,1.101) (1.18,0.216) (0.85,0.262)
        (0.91,0.748) (0.97,0.767) (1.03,0.687) (1.09,0.621)
        (1.15,0.585) (0.86,0.417) (0.92,0.508) (0.98,0.597)
        (1.04,0.880) (1.10,0.424)
    };

    \addplot[
        only marks, mark=*, mark size=1.8pt,
        draw=lotaRed!80!black, fill=lotaRed!65, opacity=0.85
    ] coordinates {
        (1.82,0.341) (1.88,0.468) (1.94,0.400) (2.00,0.491)
        (2.06,0.842) (2.12,1.194) (2.18,0.219) (1.85,0.206)
        (1.91,0.690) (1.97,0.772) (2.03,0.793) (2.09,0.647)
        (2.15,0.413) (1.86,0.383) (1.92,0.515) (1.98,0.576)
        (2.04,0.982) (2.10,0.276)
    };

    \addplot[
        only marks, mark=*, mark size=1.8pt,
        draw=lotaRed!80!black, fill=lotaRed!65, opacity=0.85
    ] coordinates {
        (2.82,0.444) (2.88,0.547) (2.94,0.468) (3.00,0.443)
        (3.06,0.909) (3.12,1.049) (3.18,0.352) (2.85,0.255)
        (2.91,0.758) (2.97,0.796) (3.03,0.617) (3.09,0.644)
        (3.15,0.530) (2.86,0.449) (2.92,0.495) (2.98,0.658)
        (3.04,0.940) (3.10,0.410)
    };

    \addplot[mark=none, line width=1.2pt, draw=lotaRed!90!black]
        coordinates {(0.74,0.572) (1.26,0.572)};
    \addplot[mark=none, line width=1.2pt, draw=lotaRed!90!black]
        coordinates {(1.74,0.567) (2.26,0.567)};
    \addplot[mark=none, line width=1.2pt, draw=lotaRed!90!black]
        coordinates {(2.74,0.598) (3.26,0.598)};

    \node[font=\small, text=lotaRed!85!black, anchor=south]
        at (axis cs:1,1.12) {18/18 wins};
    \node[font=\small, text=lotaRed!85!black, anchor=south]
        at (axis cs:2,1.20) {18/18 wins};
    \node[font=\small, text=lotaRed!85!black, anchor=south]
        at (axis cs:3,1.08) {18/18 wins};

    \end{axis}
    \end{tikzpicture}
    \end{minipage}

    \caption{Motivation and matched-control evidence for LoTA-N2N. The trace formulation identifies an interaction term separating the self-supervised MSE objective from its supervised counterpart. LoTA-N2N estimates and suppresses this interaction locally to prevent spatial cancellation; the resulting practical objective is motivated by the population supervised-risk bound derived in Section~\ref{sec4}, rather than being claimed as the bound itself. Panel~(b) reports McMaster18 per-image gains at one representative level per noise family; its mixed setting uses Poisson peak 50 with Gaussian $\sigma=20$, matching the backbone-transfer protocol rather than the $r=2$ main-table protocol. Both methods share the same 3,000-step Stage~1 checkpoint and 800-step Stage~2 budget, and LoTA-N2N uses the final trace-weight floor of 0.15.}

    \label{fig:motivation-stage2-gain}
\end{figure*}
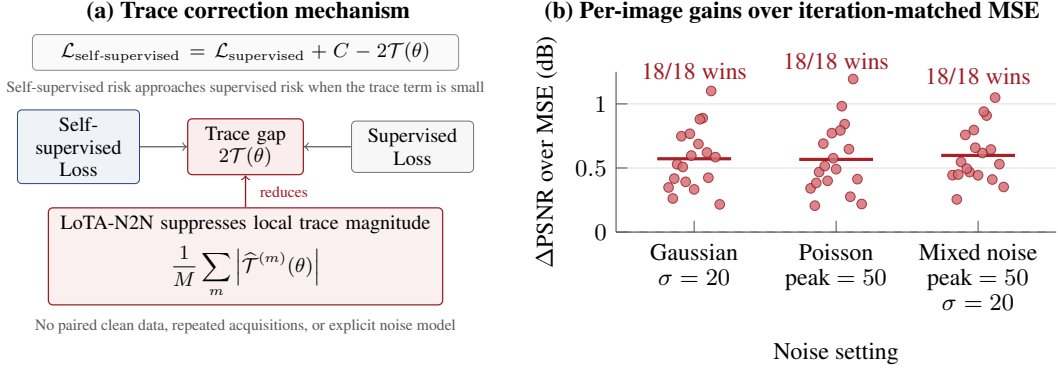

Figure~\ref{fig:motivation-stage2-gain} summarizes the trace-correction view and the iteration-matched evidence motivating the proposed adaptation.

Based on this analysis, we propose LoTA-N2N, a single-image zero-shot self-supervised denoising framework. Stage~1 constructs complementary sub-image pairs and trains a lightweight denoiser using bidirectional pair MSE with full-image consistency. A frozen copy of this denoiser is then applied to the original noisy image to produce detached clean-sub-image proxies. Stage~2 uses these proxies to estimate patch-wise residual--prediction interactions and optimizes their absolute local magnitudes together with the same MSE-based reconstruction objective.

The true local interaction magnitudes upper-bound the absolute global interaction, and we derive a population supervised-risk bound containing an explicit trace-estimation error term. The practical floor-0.15 objective uses a validation-selected, cosine-decayed trace weight and should therefore be interpreted as a bound-motivated surrogate rather than a certified instance of the theoretical upper bound. First-order discrepancy is characterized and measured diagnostically but is not included in the final training objective.

LoTA-N2N requires neither paired clean targets nor repeated noisy acquisitions and does not require the user to specify a noise distribution or re-corruption model. It nevertheless relies on sub-image redundancy and on the Stage~1 denoiser providing a useful clean-image proxy; the corresponding approximation error is explicitly reflected in our analysis. Experiments on natural, confocal, and X-ray images cover controlled synthetic and domain-specific corruptions. Iteration-matched MSE controls isolate the proposed objective from additional optimization, while assumption stress tests, paired statistical analysis, teacher-quality experiments, and gradient diagnostics evaluate whether the observed behavior is consistent with the proposed mechanism.

Our contributions are summarized as follows:
\begin{itemize}
    \item We provide a unified objective and gradient characterization of representative MSE-based self-supervised denoising methods and identify spatial cancellation as a failure mode that can remain hidden by a small global interaction.

    \item We propose LoTA-N2N, which estimates residual--prediction interactions using detached predictions from a frozen single-image teacher and suppresses their local absolute magnitudes. We prove cancellation control and derive a population supervised-risk bound with an explicit clean-estimation error term. The implemented floor-0.15 objective is a validation-selected, bound-motivated adaptation surrogate.

    \item We evaluate the proposed objective using iteration-matched controls, synthetic and cross-domain benchmarks, assumption-violating corruptions, paired statistical analysis, teacher-quality studies, and trace and gradient diagnostics. An explicit gradient-trace penalty is included only as an ablation and provides no measurable reconstruction benefit.
\end{itemize}




\section{Related Methods}
\label{sec2}

\subsection{Supervised and Repeated-Observation Denoising}

Supervised denoisers learn from aligned noisy--clean image pairs and have achieved strong performance with increasingly expressive network architectures \cite{Zhang_2017,Zhang_2018,8421285,8601375,9025693,9157160,9706178,anwar2020real,guo2019convolutional}. Their applicability is nevertheless limited when clean references are unavailable or difficult to register with noisy measurements.

Noise2Noise (N2N) \cite{lehtinen2018noise2noise} replaces each clean target with an independently corrupted observation of the same latent image. Under conditionally independent and unbiased target noise, its expected MSE differs from supervised MSE only by a parameter-independent constant. This removes the need for clean targets but still requires aligned repeated acquisitions, which may be unavailable in microscopy, medical imaging, and single-shot imaging systems.

\subsection{Single-Image Self-Supervised Denoising}

Blind-spot methods construct supervision by predicting selected pixels from their spatial context. Noise2Void \cite{krull2019noise2void} and Noise2Self \cite{batson2019noise2self} implement this principle through masking, while later methods use restricted receptive fields or asymmetric architectures \cite{laine2019highquality,lee2022apbsn,9880121}. Their theoretical justification commonly relies on conditionally zero-mean noise and independence between the held-out noise and the observed context. Spatial correlation can violate this condition, while excluding the center pixel may also remove useful information.

Noise2Same \cite{xie2020noise2same} relaxes strict $\mathcal{J}$-invariance by deriving a self-supervised upper bound on supervised denoising risk. Its objective combines full-image noisy reconstruction with a prediction-invariance discrepancy between full and masked inputs. Noise2Same therefore does not require the denoiser itself to be $\mathcal{J}$-invariant or require the noise distribution to be supplied to the algorithm, although the theoretical bound assumes zero-mean IID noise across image coordinates. It is the closest prior work to LoTA-N2N in its upper-bound motivation, but it controls a masking-based prediction discrepancy rather than estimating the surrogate residual--prediction interaction considered here.

Sub-image methods construct proxy pairs from neighboring pixels within one noisy image. Neighbor2Neighbor \cite{huang2021neighbor2neighbor} uses neighboring subsampling, while Zero-Shot Noise2Noise \cite{mansour2023zeroshot} and Noise2Fast \cite{Lequyer_2022} enable efficient per-image optimization without an external training set. These approaches are well suited to zero-shot denoising, but their proxy pairs only approximately satisfy N2N assumptions: neighboring clean pixels may differ in textured regions, and the two sub-images may retain correlated noise.

Re-corruption methods instead synthesize additional noisy observations. Noisy As Clean \cite{Xu_2020}, Noisier2Noise \cite{9156650}, Recorrupted-to-Recorrupted \cite{9577798}, and IDR \cite{zhang2022idr} construct noisier inputs or iterative targets using assumptions about the corruption process. Deep Image Prior \cite{Ulyanov_2020} and Self2Self \cite{quan2020self2self} exploit network bias, dropout, or image-specific optimization rather than paired supervision. These methods can be effective, but their behavior depends on the accuracy of the re-corruption process, architectural bias, or stochastic optimization protocol.

\subsection{Relation to LoTA-N2N}

Table~\ref{tab:trace-revisit} summarizes representative methods from the perspective developed in Section~\ref{sec:revisit}. Most classical MSE justifications attempt to make the surrogate residual--prediction interaction vanish or remain small through data construction and statistical assumptions. Noise2Same takes a related but distinct route by upper-bounding supervised risk through a full-versus-masked prediction discrepancy. LoTA-N2N estimates the interaction itself using a frozen single-image teacher and controls its local absolute magnitude to prevent spatial cancellation.

\begin{table*}[t]
    \centering
    \scriptsize
    \renewcommand{\arraystretch}{1.10}
    \setlength{\tabcolsep}{2pt}
    \begin{tabular}{l p{0.19\textwidth} p{0.27\textwidth} p{0.32\textwidth}}
        \toprule
        Method family
        & Observable construction
        & Mechanism for controlling objective mismatch
        & Principal condition or limitation \\
        \midrule

        N2N
        & Independent noisy target
        & Makes the target residual conditionally mean-zero
        & Requires registered repeated observations with conditionally independent and unbiased target noise \\

        Blind spot
        & Held-out noisy pixels
        & Removes target coordinates from the predictor so that the residual is orthogonal to the prediction
        & Relies on conditionally mean-zero, context-independent noise; spatial correlation breaks the argument \\

        Noise2Same
        & Full and masked predictions
        & Upper-bounds supervised risk using noisy reconstruction and a prediction-invariance discrepancy
        & Its bound assumes zero-mean IID coordinate noise; it uses a masking-based surrogate rather than direct interaction estimation \\

        NAC
        & Observed noise reused in the target
        & Keeps the interaction small through a weak-noise regime
        & Provides no exact zero-interaction guarantee when the observed corruption is strong \\

        R2R
        & Model-based re-corrupted pair
        & Designs the constructed input and target corruptions to be independent
        & Requires a compatible noise family and covariance or re-corruption model \\

        Sub-image pairs
        & Neighboring-pixel proxy targets
        & Combines approximate clean-neighbor similarity with noise independence
        & Texture, clean-neighbor mismatch, and correlated noise can leave a nonzero interaction \\

        \rowcolor{oursRow}\textbf{LoTA-N2N}
        & Frozen-teacher clean proxies
        & Estimates the residual--prediction interaction and penalizes its local absolute magnitude
        & Depends on teacher quality and sub-image redundancy; spatially correlated noise remains a failure boundary \\

        \bottomrule
    \end{tabular}
    \caption{Representative self-supervised denoising methods viewed through objective mismatch control. N2N, blind-spot, weak-noise, re-corruption, and sub-image methods suppress the residual--prediction interaction exactly or approximately through different assumptions. Noise2Same controls a related supervised-risk bound through prediction invariance, whereas LoTA-N2N estimates the interaction and suppresses its local magnitude during per-image adaptation.}
    \label{tab:trace-revisit}
\end{table*}

The upper-bound perspective itself is therefore not new: Noise2Same established that supervised denoising risk can be controlled without enforcing strict $\mathcal{J}$-invariance. LoTA-N2N differs in three aspects. First, it operates in the complementary sub-image, per-image adaptation setting rather than using a full-versus-masked invariance objective. Second, it constructs an explicit interaction estimate from a detached Stage~1 teacher. Third, it localizes the estimated interaction and penalizes regional absolute magnitudes, addressing spatial cancellation that is not represented by a single global discrepancy.

LoTA-N2N does not require a predefined noise distribution or re-corruption model. Its reliability instead depends on the quality of the Stage~1 clean proxy and the spatial redundancy underlying the complementary sub-image construction. The analysis in Section~\ref{sec3} characterizes this estimation error, while the correlated-noise experiments document the boundary of the complementary sub-image construction.

\section{Trace-Based Analysis}
\label{sec3}

\subsection{Objective and Gradient Discrepancies}
Let $\mathbf{x}$ be a latent clean image, $\mathbf{u}$ a self-supervised input, and $\mathbf{s}=\mathbf{x}+\mathbf{r}$ a surrogate target with residual corruption $\mathbf{r}$. Define the prediction error $\mathbf{e}_{\theta}(\mathbf{u})=f_{\theta}(\mathbf{u})-\mathbf{x}$. All squared norms and Frobenius inner products are understood as averages over their scalar entries; we omit this common normalization for readability. The supervised and self-supervised objectives are
\begin{equation}
\begin{aligned}
    \mathcal{L}_{\mathrm{sup}}(\theta)
    &=\mathbb{E}\|\mathbf{e}_{\theta}(\mathbf{u})\|_F^2,\\
    \mathcal{L}_{\mathrm{ss}}(\theta)
    &=\mathbb{E}\|f_{\theta}(\mathbf{u})-\mathbf{s}\|_F^2.
\end{aligned}
\end{equation}
Using $\|\mathbf{A}-\mathbf{B}\|_F^2=\|\mathbf{A}\|_F^2+\|\mathbf{B}\|_F^2-2\Tr(\mathbf{A}^{T}\mathbf{B})$, define
\begin{equation}
    \mathcal{T}(\theta)
    =\mathbb{E}\!\left[\left\langle
    \mathbf{r},\mathbf{e}_{\theta}(\mathbf{u})
    \right\rangle_F\right].
\end{equation}
Then
\begin{equation}
    \mathcal{L}_{\mathrm{ss}}(\theta)
    =\mathcal{L}_{\mathrm{sup}}(\theta)
    -2\mathcal{T}(\theta)+C,
\end{equation}
where $C=\mathbb{E}\|\mathbf{r}\|_F^2$ is independent of $\theta$.

\noindent\textbf{Proposition 1 (Gradient discrepancy).}
If $f_{\theta}$ is differentiable with respect to $\theta$ and differentiation
can be interchanged with expectation, then
\begin{equation}
\begin{aligned}
    \nabla_{\theta}\mathcal{L}_{\mathrm{ss}}
    -\nabla_{\theta}\mathcal{L}_{\mathrm{sup}}
    &=-2\nabla_{\theta}\mathcal{T}(\theta),\\
    \|\nabla_{\theta}\mathcal{L}_{\mathrm{ss}}
    -\nabla_{\theta}\mathcal{L}_{\mathrm{sup}}\|_2
    &=2\|\nabla_{\theta}\mathcal{T}(\theta)\|_2.
\end{aligned}
\end{equation}
\noindent\textbf{Proof.} Differentiate the exact decomposition; $C$ is independent of $\theta$.

Proposition~1 separates objective-value agreement from optimization-direction
agreement. A small trace value at a single parameter point does not by itself
imply a small gradient discrepancy. In N2N, independent zero-mean target noise
instead yields $\mathcal{T}(\theta)\equiv0$ for all $\theta$, and hence
$\nabla_{\theta}\mathcal{T}(\theta)\equiv0$. We therefore use gradient
quantities to diagnose the mechanism. The trainable objective developed below
takes a different route: local trace magnitudes alone provide an upper bound on
the inaccessible supervised risk and do not require double backpropagation.

\subsection{Revisiting Existing Methods Through the Trace Term}
\label{sec:revisit}
The preceding decomposition provides a common criterion for MSE-based
self-supervision. In particular, a convenient sufficient condition for exact
risk equivalence is
\begin{equation}
    \mathbb{E}[\mathbf{r}\mid\mathbf{u},\mathbf{x}]=\mathbf{0}.
    \label{eq:conditional-trace-zero}
\end{equation}
Indeed, conditioning on $(\mathbf{u},\mathbf{x})$ gives
\begin{equation}
\begin{aligned}
    \mathcal{T}(\theta)
    =\mathbb{E}_{\mathbf{u},\mathbf{x}}
    \left\langle
    \mathbb{E}[\mathbf{r}\mid\mathbf{u},\mathbf{x}],
    f_{\theta}(\mathbf{u})-\mathbf{x}
    \right\rangle_F
    =0
\end{aligned}
\end{equation}
for every $\theta$. Consequently, $\nabla_{\theta}\mathcal{T}(\theta)=0$
and the self-supervised and supervised gradients coincide. Equation
\eqref{eq:conditional-trace-zero} is sufficient rather than necessary; the
more general requirement is
$\mathbb{E}\langle\mathbf{r},\mathbf{e}_{\theta}(\mathbf{u})\rangle_F=0$
for all $\theta$. This distinction reveals that several apparently different
methods suppress the same interaction through different data constructions or
statistical assumptions. Table~\ref{tab:trace-revisit} summarizes the common
mechanism and the distinct sufficient conditions.

\noindent\textbf{Independent noisy targets.}
For N2N, $\mathbf{u}=\mathbf{x}+\mathbf{n}$,
$\mathbf{s}=\mathbf{x}+\mathbf{n}'$, and $\mathbf{r}=\mathbf{n}'$.
If $\mathbb{E}[\mathbf{n}'\mid\mathbf{x}]=0$ and
$\mathbf{n}'\perp\!\!\!\perp\mathbf{n}\mid\mathbf{x}$, then
$\mathbb{E}[\mathbf{n}'\mid\mathbf{u},\mathbf{x}]=0$, so
$\mathcal{T}(\theta)\equiv0$. Gaussianity and equal noise variances are not
essential to this argument; conditional unbiasedness and independence are the
key requirements. The practical restriction is the need for aligned repeated
observations whose target corruption is independent of the input corruption.

\noindent\textbf{Blind-spot prediction.}
For a held-out coordinate set $J$, blind-spot methods predict
$\mathbf{y}_{J}=\mathbf{x}_{J}+\mathbf{n}_{J}$ using only
$\mathbf{y}_{J^c}$. Their trace interaction is
\begin{equation}
    \mathbb{E}\left\langle
    \mathbf{n}_{J},
    f_{\theta,J}(\mathbf{y}_{J^c})-\mathbf{x}_{J}
    \right\rangle_F.
\end{equation}
The $\mathcal{J}$-invariant construction makes this term zero when the held-out
noise is conditionally mean-zero and independent of the observed-coordinate
noise given the clean image \cite{batson2019noise2self}. Thus, masking is not
merely an architectural device: in the unified view, it is a mechanism for
enforcing trace orthogonality. Spatially correlated noise violates this
mechanism.

\noindent\textbf{Weak-noise targets.}
NAC forms $\mathbf{u}=\mathbf{x}+\mathbf{n}_{o}+\mathbf{n}_{s}$ and uses
$\mathbf{s}=\mathbf{x}+\mathbf{n}_{o}$ as the target. Here the target residual
$\mathbf{n}_{o}$ is also contained in the input, so
Eq.~\eqref{eq:conditional-trace-zero} generally does not hold. Instead,
Cauchy--Schwarz gives
\begin{equation}
\begin{aligned}
    |\mathcal{T}(\theta)|
    &\le
    \sqrt{\mathbb{E}\|\mathbf{n}_{o}\|_F^2}
    \sqrt{\mathbb{E}\|\mathbf{e}_{\theta}(\mathbf{u})\|_F^2},\\
    \|\nabla_{\theta}\mathcal{T}(\theta)\|_2
    &\le
    \sqrt{\mathbb{E}\|\mathbf{n}_{o}\|_F^2}
    \sqrt{\mathbb{E}\|\mathbf{J}_{\theta}f_{\theta}(\mathbf{u})\|_F^2},
\end{aligned}
\end{equation}
where $\mathbf{J}_{\theta}f_{\theta}(\mathbf{u})$ is the Jacobian of the
network output with respect to its parameters.
The weak-noise assumption therefore has a direct interpretation: it controls
the trace value and its gradient only approximately through the energy of the
reused target noise.

\noindent\textbf{Model-based re-corruption.}
R2R starts from $\mathbf{y}=\mathbf{x}+\mathbf{n}$ and constructs
\begin{equation}
    \widehat{\mathbf{y}}=\mathbf{y}+\mathbf{A}\mathbf{z},\qquad
    \widetilde{\mathbf{y}}=\mathbf{y}-\mathbf{B}\mathbf{z}.
\end{equation}
For $\mathbf{n}\mid\mathbf{x}\sim
\mathcal{N}(0,\boldsymbol{\Sigma}_{\mathbf{x}})$,
$\mathbf{z}\sim\mathcal{N}(0,\mathbf{I})$, and
$\mathbf{A}\mathbf{B}^{T}=\boldsymbol{\Sigma}_{\mathbf{x}}$, the constructed
noises satisfy
$\operatorname{Cov}(\widehat{\mathbf{n}},\widetilde{\mathbf{n}}\mid\mathbf{x})
=\boldsymbol{\Sigma}_{\mathbf{x}}-\mathbf{A}\mathbf{B}^{T}=0$. Joint
Gaussianity then turns zero covariance into conditional independence, again
making the trace term vanish. The cost of this exact cancellation is a
specified noise family and a compatible covariance model.

\noindent\textbf{Sub-image pairs and the role of LoTA-N2N.}
For neighboring sub-images, taking $\mathbf{x}_1$ as the clean target gives
$\mathbf{r}_{1\rightarrow2}=\mathbf{y}_2-\mathbf{x}_1
=(\mathbf{x}_2-\mathbf{x}_1)+\mathbf{n}_2$. Pixelwise independent,
zero-mean noise can remove the noise contribution in expectation, while local
similarity is used to keep $\mathbf{x}_2-\mathbf{x}_1$ small. Neither property
is guaranteed for textured content or structured noise, so a residual trace
interaction can remain.

This taxonomy is the conceptual departure of LoTA-N2N. Existing objectives
control $\mathcal{T}$ indirectly through repeated acquisition, masking,
weak-noise assumptions, or a prescribed re-corruption model. LoTA-N2N instead
uses the Stage~1 denoiser to estimate the residual interaction and directly
reduces its local magnitude. The method is therefore not assumption-free: it
replaces exact statistical cancellation with an approximation whose error is
governed by the clean-estimation error. First-order mismatch remains useful for
mechanism analysis, but it is not directly regularized by the final method. The next
section quantifies this trade-off and motivates the two-stage design.

\section{LoTA-N2N}
\label{sec4}

\begin{figure*}[t]
    \centering
    \safeincludegraphics[width=1\linewidth]{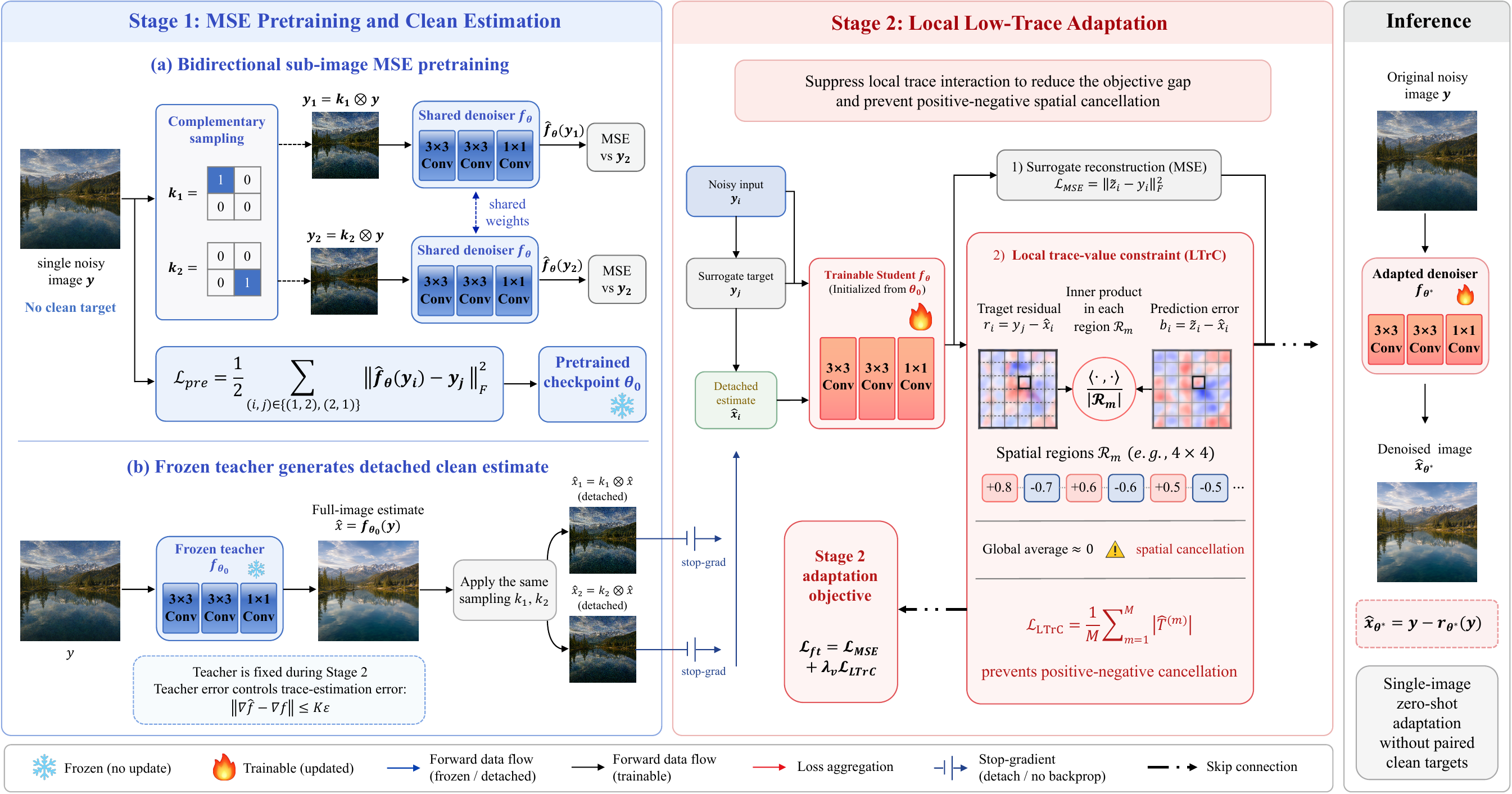}
    \caption{Pipeline of LoTA-N2N. Stage~1 pretrains the denoiser with
    bidirectional sub-image MSE. The frozen checkpoint then produces detached
    clean-sub-image estimates. Stage~2 fine-tunes the trainable denoiser using
    only bidirectional MSE and the patch-wise local trace correction
    $\mathcal{L}_{\mathrm{LTrC}}$.}
    \label{fig:pipeline}
\end{figure*}

The overall pipeline is shown in Figure~\ref{fig:pipeline}. Given one noisy image $\mathbf{y}$, two correlated sub-images are generated by
\begin{equation}
    \mathbf{y}_1=\mathbf{k}_1\otimes\mathbf{y},\qquad
    \mathbf{y}_2=\mathbf{k}_2\otimes\mathbf{y},
\end{equation}
where $\mathbf{k}_1$ and $\mathbf{k}_2$ are $2\times2$ sampling kernels. Because the resulting pair is not guaranteed to satisfy N2N independence, direct MSE training can retain a nonzero trace interaction.

\subsection{Stage 1: MSE Pretraining}
The first stage minimizes
\begin{equation}
    \mathcal{L}_{\mathrm{pre}}(\theta)
    =\mathbb{E}\|f_{\theta}(\mathbf{y}_1)-\mathbf{y}_2\|_F^2.
\end{equation}
Let $\theta_0$ denote the pretrained parameters. During Stage~2, $\theta_0$ is frozen and the clean sub-image estimates are detached:
\begin{equation}
    \widehat{\mathbf{x}}_i
    =\operatorname{sg}\!\left(
    \mathbf{k}_i\otimes f_{\theta_0}(\mathbf{y})
    \right),\quad i\in\{1,2\},
\end{equation}
where $\operatorname{sg}(\cdot)$ denotes stop-gradient.

\subsection{Reliability of the Estimated Trace}
For direction $i\rightarrow j$, define $\mathbf{e}_i=\widehat{\mathbf{x}}_i-\mathbf{x}_i$, $\mathbf{p}_{\theta,i}=f_{\theta}(\mathbf{y}_i)-\mathbf{x}_i$, and $\mathbf{q}_{i\rightarrow j}=\mathbf{y}_j-\mathbf{x}_i$. Let
\begin{equation}
\begin{aligned}
    \mathcal{T}_{i\rightarrow j}
    &=\mathbb{E}\langle\mathbf{q}_{i\rightarrow j},
    \mathbf{p}_{\theta,i}\rangle_F,\\
    \widehat{\mathcal{T}}_{i\rightarrow j}
    &=\mathbb{E}\langle
    \mathbf{y}_j-\widehat{\mathbf{x}}_i,
    f_{\theta}(\mathbf{y}_i)-\widehat{\mathbf{x}}_i
    \rangle_F.
\end{aligned}
\end{equation}

\noindent\textbf{Proposition 2 (Estimation error).}
If $\mathbb{E}\|\mathbf{e}_i\|_F^2\le\epsilon^2$, $\mathbb{E}\|\mathbf{p}_{\theta,i}\|_F^2\le P^2$, and $\mathbb{E}\|\mathbf{q}_{i\rightarrow j}\|_F^2\le Q^2$, then
\begin{equation}
    |\widehat{\mathcal{T}}_{i\rightarrow j}
    -\mathcal{T}_{i\rightarrow j}|
    \le\epsilon(P+Q)+\epsilon^2.
\end{equation}
\noindent\textbf{Proof.} Expanding with $\widehat{\mathbf{x}}_i=\mathbf{x}_i+\mathbf{e}_i$ gives
\begin{equation}
\begin{aligned}
    \widehat{\mathcal{T}}-\mathcal{T}
    =\mathbb{E}[&-\langle\mathbf{e}_i,\mathbf{p}_{\theta,i}\rangle_F
    -\langle\mathbf{q}_{i\rightarrow j},\mathbf{e}_i\rangle_F
    +\|\mathbf{e}_i\|_F^2].
\end{aligned}
\end{equation}
The result follows from the triangle and Cauchy--Schwarz inequalities.

If the denoiser Jacobian $\mathbf{J}_{\theta,i}$ satisfies $\|\mathbf{J}_{\theta,i}\|_{\mathrm{op}}\le K$, stop-gradient further gives
\begin{equation}
    \|\nabla_{\theta}\widehat{\mathcal{T}}_{i\rightarrow j}
    -\nabla_{\theta}\mathcal{T}_{i\rightarrow j}\|_2
    \le K\epsilon.
\end{equation}
Thus, improving the Stage~1 estimate tightens the value-level risk surrogate.
The gradient bound is retained only to interpret the diagnostic measurements in
Section~\ref{sec:mechanism}.

\subsection{Local Low-Trace Adaptation}
We partition each sub-image into $M$ equal-sized spatial regions
$\{\Omega_m\}_{m=1}^{M}$. For direction $i\rightarrow j$, define the true and
estimated local interactions as
\begin{equation}
\begin{aligned}
    \mathcal{T}_{i\rightarrow j}^{(m)}
    &=
    \mathbb{E}\!\left[
    \frac{1}{|\Omega_m|}
    \left\langle
    (\mathbf{q}_{i\rightarrow j})_{\Omega_m},
    (\mathbf{p}_{\theta,i})_{\Omega_m}
    \right\rangle_F
    \right],\\
    \widehat{\mathcal{T}}_{i\rightarrow j}^{(m)}
    &=
    \mathbb{E}\!\left[
    \frac{1}{|\Omega_m|}
    \left\langle
    (\mathbf{y}_j-\widehat{\mathbf{x}}_i)_{\Omega_m},
    (f_{\theta}(\mathbf{y}_i)-\widehat{\mathbf{x}}_i)_{\Omega_m}
    \right\rangle_F
    \right].
\end{aligned}
\end{equation}
The expectations are implemented empirically using the sampled sub-image pairs
and augmentations at each update. For equal-sized regions,
$\mathcal{T}_{i\rightarrow j}=M^{-1}\sum_m
\mathcal{T}_{i\rightarrow j}^{(m)}$, with the analogous identity for the
estimated interaction. The bidirectional local trace loss is
\begin{equation}
    \mathcal{L}_{\mathrm{LTrC}}
    =\frac{1}{2M}
    \sum_{(i,j)\in\{(1,2),(2,1)\}}
    \sum_{m=1}^{M}
    |\widehat{\mathcal{T}}_{i\rightarrow j}^{(m)}|.
\end{equation}

\noindent\textbf{Proposition 3 (Cancellation control).}
For either direction,
\begin{equation}
    \left|\frac{1}{M}\sum_m
    \widehat{\mathcal{T}}_{i\rightarrow j}^{(m)}\right|
    \le\frac{1}{M}\sum_m
    |\widehat{\mathcal{T}}_{i\rightarrow j}^{(m)}|.
\end{equation}
Therefore, the local loss upper-bounds the absolute global interaction and
prevents positive and negative regions from cancelling.

To connect this value-level control directly to the inaccessible supervised
objective, define
\begin{equation}
\begin{aligned}
    \mathcal{L}_{\mathrm{MSE}}
    &=
    \frac{1}{2}
    \sum_{(i,j)\in\{(1,2),(2,1)\}}
    \mathbb{E}\|f_{\theta}(\mathbf{y}_i)-\mathbf{y}_j\|_F^2,\\
    \mathcal{L}_{\mathrm{sup}}^{\mathrm{bi}}
    &=
    \frac{1}{2}
    \sum_{i\in\{1,2\}}
    \mathbb{E}\|f_{\theta}(\mathbf{y}_i)-\mathbf{x}_i\|_F^2,
\end{aligned}
\end{equation}
with the corresponding average residual-energy constant
\begin{equation}
    C_{\mathrm{bi}}
    =
    \frac{1}{2}
    \sum_{(i,j)\in\{(1,2),(2,1)\}}
    \mathbb{E}\|\mathbf{y}_j-\mathbf{x}_i\|_F^2.
\end{equation}
We also define the inaccessible true local-trace magnitude and its estimation
error:
\begin{equation}
\begin{aligned}
    \mathcal{L}_{\mathrm{LTrC}}^{\star}
    &=
    \frac{1}{2M}
    \sum_{(i,j)\in\{(1,2),(2,1)\}}
    \sum_{m=1}^{M}
    |\mathcal{T}_{i\rightarrow j}^{(m)}|,\\
    \Delta_{\mathcal{T}}
    &=
    \frac{1}{2M}
    \sum_{(i,j)\in\{(1,2),(2,1)\}}
    \sum_{m=1}^{M}
    |\widehat{\mathcal{T}}_{i\rightarrow j}^{(m)}
    -\mathcal{T}_{i\rightarrow j}^{(m)}|.
\end{aligned}
\end{equation}

\noindent\textbf{Theorem 1 (Supervised-risk upper bound).}
For the bidirectional sub-image objective,
\begin{equation}
\begin{aligned}
    \mathcal{L}_{\mathrm{sup}}^{\mathrm{bi}}(\theta)
    &\le
    \mathcal{L}_{\mathrm{MSE}}(\theta)
    -C_{\mathrm{bi}}
    +2\mathcal{L}_{\mathrm{LTrC}}^{\star}(\theta)\\
    &\le
    \mathcal{L}_{\mathrm{MSE}}(\theta)
    -C_{\mathrm{bi}}
    +2\mathcal{L}_{\mathrm{LTrC}}(\theta)
    +2\Delta_{\mathcal{T}}(\theta).
\end{aligned}
\end{equation}
If the assumptions of Proposition~2 hold patch-wise with common bounds
$\epsilon$, $P$, and $Q$, then
\begin{equation}
    \Delta_{\mathcal{T}}
    \le \epsilon(P+Q)+\epsilon^2.
\end{equation}

\noindent\textbf{Proof.}
Apply the exact decomposition to each direction:
\begin{equation}
    \mathcal{L}_{\mathrm{sup}}^{i}
    =
    \mathcal{L}_{\mathrm{ss}}^{i\rightarrow j}
    -C_{i\rightarrow j}
    +\frac{2}{M}\sum_m
    \mathcal{T}_{i\rightarrow j}^{(m)}.
\end{equation}
The first inequality follows from
$\sum_m\mathcal{T}^{(m)}\le\sum_m|\mathcal{T}^{(m)}|$ and averaging the two
directions. The second follows from
$|\mathcal{T}^{(m)}|\le
|\widehat{\mathcal{T}}^{(m)}|+
|\widehat{\mathcal{T}}^{(m)}-\mathcal{T}^{(m)}|$.
The final bound is Proposition~2 applied to each normalized patch and then
averaged.

Theorem~1 is the main justification for using LTrC alone: the correction term
required to upper-bound supervised risk depends only on local trace values, not
on their parameter gradients. The coefficient $2$ has a direct population-risk
interpretation under the normalization above; the practical weight
$\lambda_{\mathrm{v}}$ absorbs finite-sample scaling and is selected on the
validation set.

Proposition~1 remains important for interpretation: without additional
regularity, no value-only penalty can universally guarantee gradient equality.
Accordingly, the final method claims the supervised-risk bound above, while
gradient discrepancy is evaluated only as a diagnostic in
Section~\ref{sec:mechanism}.

The final fine-tuning objective is therefore
\begin{equation}
    \mathcal{L}_{\mathrm{ft}}
    =
    \mathcal{L}_{\mathrm{MSE}}
    +\lambda_{\mathrm{v}}\mathcal{L}_{\mathrm{LTrC}}.
\end{equation}
The trace weight follows a cosine schedule. Residual enhancement parameterizes
the denoised output as
$f_{\theta}(\mathbf{y})=\mathbf{y}-r_{\theta}(\mathbf{y})$, where
$r_{\theta}$ predicts the noise residual.

\section{Experiments}
\label{sec5}

\subsection{Datasets and Evaluation Protocol}

\begin{table*}[t]
    \centering
    
    \scriptsize
    \renewcommand{\arraystretch}{1.0}
    \setlength{\tabcolsep}{1.5pt}
    
    \resizebox{\textwidth}{!}{%
    \begin{tabular}{l | c | ccc | ccc | ccc | ccc}
        \toprule
        \multirow{2}{*}{Method}
        & \multirow{2}{*}{Mode}
        & \multicolumn{3}{c|}{Kodak24}
        & \multicolumn{3}{c|}{McMaster18}
        & \multicolumn{3}{c|}{Set14}
        & \multicolumn{3}{c}{Set5} \\
        \cmidrule(lr){3-5}
        \cmidrule(lr){6-8}
        \cmidrule(lr){9-11}
        \cmidrule(lr){12-14}
        &
        & \multicolumn{3}{c|}{Noise level}
        & \multicolumn{3}{c|}{Noise level}
        & \multicolumn{3}{c|}{Noise level}
        & \multicolumn{3}{c}{Noise level} \\
        
        \midrule
        \textit{Gaussian noise} 
        &
        & $\sigma=10$ & $\sigma=15$ & $\sigma=20$
        & $\sigma=10$ & $\sigma=15$ & $\sigma=20$
        & $\sigma=10$ & $\sigma=15$ & $\sigma=20$
        & $\sigma=10$ & $\sigma=15$ & $\sigma=20$ \\
        \midrule

        DnCNN
        & Offline & \textbf{35.33} & \textbf{33.17} & \textbf{31.50}
        & \textbf{34.48} & \textbf{32.89} & \textbf{31.51}
        & \textbf{33.65} & \textbf{31.99} & \textbf{30.68}
        & \textbf{35.01} & \textbf{33.26} & \textbf{31.83} \\
        
        N2N
        & Offline & 34.86 & 32.88 & 31.46 & 33.57 & 32.22 & 31.06
        & 33.07 & 31.63 & 30.49
        & 34.61 & 33.07 & 31.81 \\ \hline
CBM3D
        & Online & 33.50 & 31.30 & 29.83 & 34.49 & 32.18 & 30.48 & 32.92 & 30.74 & 29.22
        & \textbf{34.38} & \textbf{32.85} & \textbf{31.67} \\
        
        DIP2000
        & Online & 33.13 & 31.13 & 29.69 & 33.65 & 31.86 & 30.34 & 29.91 & 29.26 & 28.26
        & 30.50 & 29.67 & 28.83 \\
        
        Noise2Self
        & Online & 28.80 & 28.23 & 27.44 & 30.46 & 29.64 & 28.62
        & 29.58 & 28.37 & 28.21
        & 30.52 & 29.23 & 27.95 \\
        
        Noise2Fast
        & Online & 32.22 & 30.78 & 29.63 & 33.89 & 32.10 & 30.64 & 31.49 & 30.08 & 28.93
        & 33.11 & 31.58 & 30.33 \\
        
        ZSN2N
        & Online & 33.91 & 31.98 & 30.43 & 34.19 & 32.00 & 30.31 & 33.00 & 31.02 & 29.55
        & 33.82 & 31.88 & 30.32 \\
        
        \rowcolor{oursRow}\textbf{LoTA-N2N}
        & Online
        & \textbf{34.46}
        & \textbf{32.39}
        & \textbf{30.87}
        & \textbf{34.58}
        & \textbf{32.41}
        & \textbf{30.77}
        & \textbf{33.12}
        & \textbf{31.23}
        & \textbf{29.84}
        & 34.03
        & 32.16
        & 30.67 \\
        
        \midrule
        \textit{Poisson noise} 
        &
        & $\lambda=60$ & $\lambda=50$ & $\lambda=40$
        & $\lambda=60$ & $\lambda=50$ & $\lambda=40$
        & $\lambda=60$ & $\lambda=50$ & $\lambda=40$
        & $\lambda=60$ & $\lambda=50$ & $\lambda=40$ \\
        \midrule

        DnCNN
        & Offline & \textbf{31.80} & \textbf{31.35} & \textbf{30.68}
        & 24.03 & 23.89 & 23.67
        & 26.88 & 26.61 & 26.20
        & 23.39 & 23.26 & 23.06 \\
        
        N2N
        & Offline & 28.25 & 28.83 & 29.37
        & \textbf{28.56} & \textbf{29.18} & \textbf{29.80}
        & \textbf{27.09} & \textbf{27.81} & \textbf{28.33}
        & \textbf{27.98} & \textbf{28.38} & \textbf{29.37} \\ \hline

CBM3D
        & Online & 28.33 & 28.26 & 28.08 & 29.33 & 29.21 & 28.97
        & 29.37 & 28.79 & 28.21
        & 28.74 & 28.10 & 27.38 \\
        
        DIP2000
        & Online & 29.11 & 28.62 & 28.04 & 30.29 & 29.78 & 29.33
        & 26.12 & 25.83 & 25.48
        & 28.70 & 28.26 & 27.65 \\
        
        Noise2Self
        & Online & 27.08 & 26.77 & 26.67 & 29.03 & 29.00 & 28.31
        & 27.75 & 27.24 & 27.00
        & 27.93 & 26.94 & 26.56 \\
        
        Noise2Fast
        & Online & 29.29 & 28.87 & 28.37 & 31.01 & 30.54 & 29.98
        & 28.61 & 28.20 & 27.71
        & 30.41 & 29.98 & 29.42 \\
        
        ZSN2N
        & Online & 30.36 & \textbf{29.93} & 29.28 & 30.80 & 30.47 & 29.86
        & 29.09 & 28.63 & 27.98
        & 30.36 & 29.82 & 29.25 \\
        
        \rowcolor{oursRow}\textbf{LoTA-N2N}
        & Online
        & \textbf{30.45}
        & \textbf{29.93}
        & \textbf{29.29}
        & \textbf{31.35}
        & \textbf{30.84}
        & \textbf{30.22}
        & \textbf{29.43}
        & \textbf{28.92}
        & \textbf{28.31}
        & \textbf{30.67}
        & \textbf{30.17}
        & \textbf{29.54} \\

        \midrule
        \textit{Mixed noise}
        &
        & \makecell{$p=60$\\ $r=2$}
        & \makecell{$p=50$\\ $r=2$}
        & \makecell{$p=40$\\ $r=2$}
        & \makecell{$p=60$\\ $r=2$}
        & \makecell{$p=50$\\ $r=2$}
        & \makecell{$p=40$\\ $r=2$}
        & \makecell{$p=60$\\ $r=2$}
        & \makecell{$p=50$\\ $r=2$}
        & \makecell{$p=40$\\ $r=2$}
        & \makecell{$p=60$\\ $r=2$}
        & \makecell{$p=50$\\ $r=2$}
        & \makecell{$p=40$\\ $r=2$} \\
        \midrule

        DnCNN
        & Offline
        & \textbf{31.72} & \textbf{31.24} & \textbf{30.53}
        & 21.80 & 21.72 & 21.57
        & 25.94 & 25.68 & 25.31
        & 20.76 & 20.68 & 20.53 \\
        
        N2N
        & Offline
        & 29.08 & 29.53 & 29.59
        & \textbf{29.46} & \textbf{29.98} & \textbf{30.28}
        & \textbf{28.10} & \textbf{28.60} & \textbf{28.74}
        & \textbf{28.98} & \textbf{29.42} & \textbf{30.06} \\ \hline

CBM3D
        & Online
        & 30.37 & 29.85 & 29.22
        & 29.00 & 28.35 & 27.59
        & 29.24 & 28.71 & 28.19
        & 28.80 & 28.12 & 27.32 \\
        
        DIP2000
        & Online
        & 26.63 & 26.35 & 25.97
        & 28.15 & 27.97 & 27.56
        & 26.11 & 25.81 & 25.49
        & 28.55 & 28.19 & 27.59 \\
        
        Noise2Self
        & Online
        & 28.24 & 28.17 & 27.79
        & 30.10 & 29.86 & 29.63
        & 27.89 & 27.40 & 26.92
        & 27.95 & 27.51 & 26.26 \\
        
        Noise2Fast
        & Online
        & 29.38 & 28.97 & 28.45
        & 31.14 & 30.70 & \textbf{30.14}
        & 28.58 & 28.17 & 27.67
        & 30.40 & 29.99 & 29.41 \\

        ZSN2N
        & Online & 30.03 & 29.50 & 28.85 & 30.73 & 30.23 & 29.58
        & 29.08 & 28.57 & 27.95
        & 30.33 & 29.84 & 29.24 \\

        \rowcolor{oursRow}\textbf{LoTA-N2N}
        & Online
        & \textbf{30.44}
        & \textbf{29.92}
        & \textbf{29.27}
        & \textbf{31.27}
        & \textbf{30.76}
        & \textbf{30.14}
        & \textbf{29.38}
        & \textbf{28.89}
        & \textbf{28.28}
        & \textbf{30.62}
        & \textbf{30.17}
        & \textbf{29.52} \\
        
        \bottomrule
    \end{tabular}%
    }
    \caption{
        Quantitative comparison of different denoising methods on Kodak24, McMaster18, Set14, and Set5.
        The evaluation metric is PSNR (dB). The best offline result and the best online result in each setting are highlighted separately in bold; ties at the reported precision are all highlighted.
        In the mixed-noise block, $p$ denotes the Poisson peak value and $r$ denotes the Gaussian read-noise standard deviation on the 0--255 intensity scale.
        DnCNN and N2N are randomly initialized dataset-trained methods: models trained on Kodak24 are evaluated on McMaster18, Set14, and Set5, while models trained on McMaster18 are evaluated on Kodak24. No official pretrained weights are used. All newly reproduced entries use the same saved noisy inputs as LoTA-N2N.
    }
    \label{tab:kodak-mcmaster}
\end{table*}

\begin{table*}[t]
    \centering
    
    \small
    \renewcommand{\arraystretch}{1.0}
    \setlength{\tabcolsep}{2pt}
    
    \begin{tabular}{l | cc | cc | cc | cc}
        \toprule
        \multirow{3}{*}{Method}
        & \multicolumn{4}{c|}{Confocal ($500\times500$)}
        & \multicolumn{4}{c}{X-ray ($800\times800$)} \\
        \cmidrule(lr){2-5}
        \cmidrule(lr){6-9}
        
        & \multicolumn{2}{c|}{Gaussian noise}
        & \multicolumn{2}{c|}{Poisson noise}
        & \multicolumn{2}{c|}{Gaussian noise}
        & \multicolumn{2}{c}{Poisson noise} \\
        \cmidrule(lr){2-3}
        \cmidrule(lr){4-5}
        \cmidrule(lr){6-7}
        \cmidrule(lr){8-9}
        
        & $\sigma=5$
        & $\sigma=10$
        & $\lambda=60$
        & $\lambda=50$
        & $\sigma=5$
        & $\sigma=10$
        & $\lambda=60$
        & $\lambda=50$ \\
        
        \midrule
        
        CBM3D
        & 42.47
        & 38.28
        & 36.87
        & 36.70
        & 41.30
        & 38.60
        & 35.57
        & 35.25 \\
        
        DIP2000
        & 38.98
        & 37.11
        & 37.20
        & 36.84
        & 36.21
        & 35.95
        & 35.53
        & 35.33 \\
        
        Noise2Fast
        & 41.49
        & 38.98
        & 38.52
        & 38.25
        & 40.83
        & 38.40
        & 35.32
        & 34.84 \\
        
        ZSN2N
        & 44.13
        & 39.01
        & 39.81
        & 39.33
        & 42.04
        & 39.06
        & 35.79
        & 35.31 \\
        
        \rowcolor{oursRow}\textbf{LoTA-N2N}
        & \textbf{44.21}
        & \textbf{39.26}
        & \textbf{40.17}
        & \textbf{39.65}
        & \textbf{42.96}
        & \textbf{39.74}
        & \textbf{35.81}
        & \textbf{35.35} \\
        
        \bottomrule
    \end{tabular}
    \caption{
        Quantitative comparison of different denoising methods on the Confocal and X-ray datasets.
        The evaluation metric is PSNR (dB), and the best result in each setting is highlighted in bold.
        The LoTA-N2N row uses the final floor-0.15 configuration.
    }
    \label{tab:confocal-xray}
\end{table*}

We first evaluate controlled synthetic corruption on Kodak24, McMaster18
\cite{McMaster18}, Set14 \cite{Set14}, and Set5 \cite{bevilacqua2012low}. Cross-domain
evaluation uses confocal images from FMD \cite{zhang2019poissongaussian} and
pediatric chest X-ray images \cite{xray}. Kodak24 and McMaster18 are evaluated
at $500\times500$, while the selected X-ray images are evaluated at
$800\times800$. Each noisy image is adapted independently, and no clean target
is exposed to the optimization procedure.

The standard synthetic evaluation includes additive Gaussian,
signal-dependent Poisson, and mixed Poisson--Gaussian noise. Two mixed-noise
protocols are reported and are never compared across protocols: the main
benchmark, paired statistics, and Kodak gain-distribution figure use Gaussian
read noise $r=2$ on the 0--255 scale, whereas the introductory McMaster18
matched-control plot and the backbone-transfer study use the stronger setting
with Poisson peak 50 and Gaussian $\sigma=20$. Section~\ref{sec:stress-test}
further studies spatially varying Gaussian and a separate mixed-noise shift,
while Section~\ref{sec:limitations} reports a correlated-Gaussian stress
check. Inputs are represented in $[0,1]$ and clipped to this range after
corruption. Reconstruction quality is measured by PSNR and SSIM \cite{ssim}.

Aggregation protocols are stated with each table. In the selected noise-shift
experiment, every cell averages five independently generated noise realizations
per image using a fixed optimization seed of 2027, and all methods receive the
same saved noisy inputs. Paired 95\% confidence intervals use 10,000 image-level
bootstrap resamples after first averaging the five realizations for each image.
This matched protocol is necessary because several improvements are smaller
than $0.2$ dB.

\subsection{Implementation Details}
The denoising backbone is a lightweight three-layer residual CNN with 48 hidden
channels. The convolution kernels are $3\times3$, $3\times3$, and $1\times1$,
and Leaky-ReLU follows the first two convolutions. Stage~1 uses bidirectional
sub-image MSE and Adam for 3,000 steps with an initial learning rate of
$10^{-3}$; a StepLR schedule halves the rate every 1,000 steps. Stage~2
initializes the trainable denoiser from the frozen Stage~1 checkpoint and uses
Adam for 800 steps with a cosine learning-rate schedule from $10^{-4}$ to
$10^{-5}$. The local region width is 32 pixels,
$\lambda_{\mathrm{v}}=8$, and the trace weight follows a cosine decay to
15\% of its initial value. Unless explicitly stated otherwise, all LoTA-N2N
results use this floor-0.15 configuration. The final method optimizes only
$\mathcal{L}_{\mathrm{MSE}}+\lambda_{\mathrm{v}}
\mathcal{L}_{\mathrm{LTrC}}$. A gradient-trace penalty is evaluated only in
the mechanism ablation because the Set12 sweep showed no PSNR benefit and it
requires double backpropagation. Hyperparameters and checkpoint selection are
fixed on Set12 before evaluation on the reported test sets.

\subsection{Comparison with Existing Methods}
We compare LoTA-N2N with representative classical, supervised,
self-supervised, and zero-shot denoisers. Table~\ref{tab:kodak-mcmaster}
includes CBM3D \cite{bm3d}, DnCNN \cite{Zhang_2017}, N2N
\cite{lehtinen2018noise2noise}, DIP \cite{Ulyanov_2020}, Noise2Self
\cite{batson2019noise2self}, Noise2Fast \cite{Lequyer_2022}, and ZSN2N
\cite{mansour2023zeroshot}. DnCNN and N2N are offline, dataset-trained
references, whereas the remaining methods operate without dataset-level
training on the reported test images. Self2Self \cite{quan2020self2self} is
included separately in the step-matched noise-shift and runtime comparisons,
where its reimplementation status is stated explicitly.

\begin{figure*}[!t]
    \centering
    \safeincludegraphics[width=\textwidth]{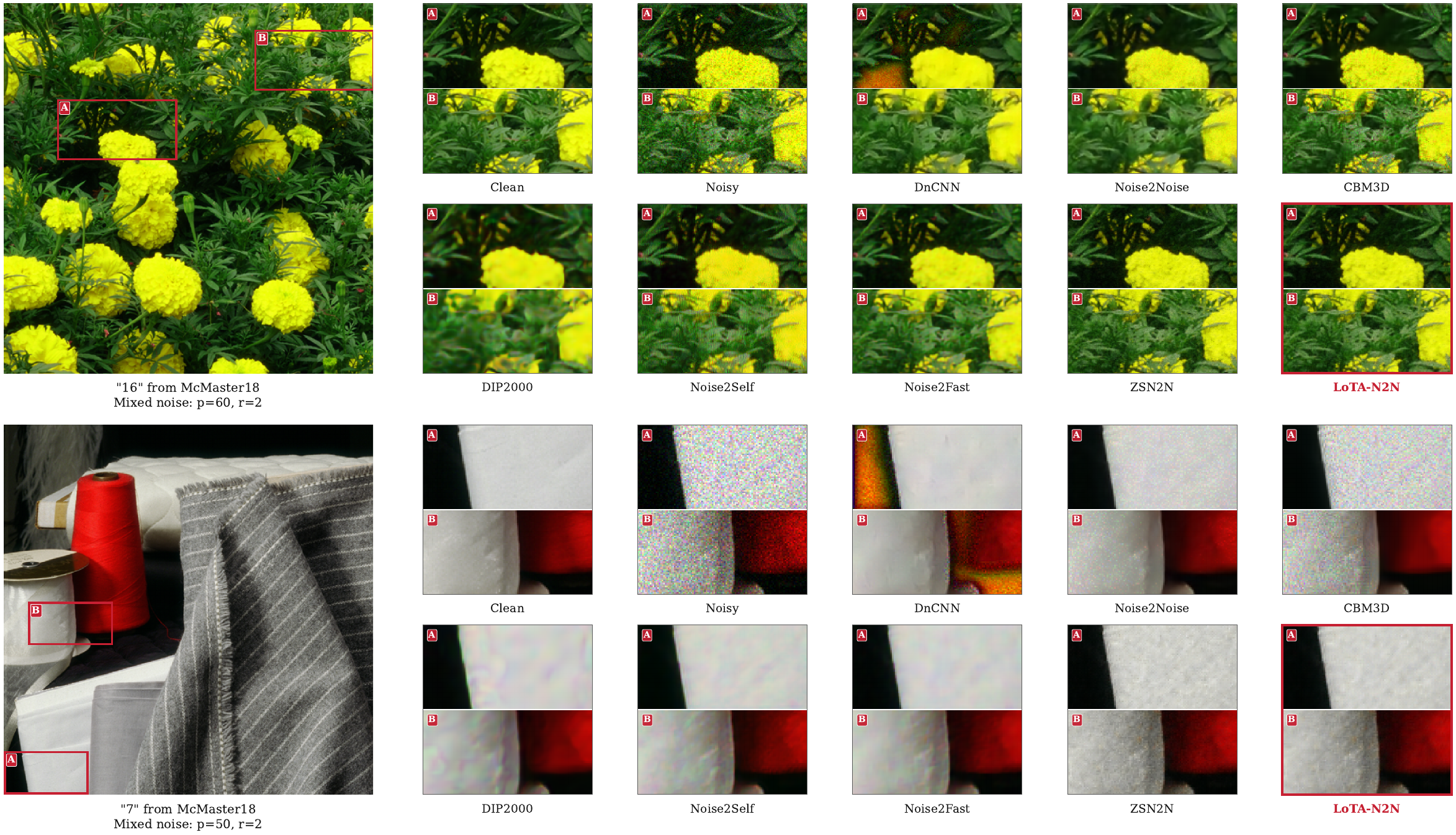}
    \caption{Qualitative comparison on natural images. Enlarged regions highlight fine structures and texture-rich details. LoTA-N2N effectively suppresses noise while preserving edges and subtle textures without introducing noticeable artifacts.}
    \label{fig:visual-natural}
\end{figure*}

\begin{figure*}[!t]
    \centering
    \safeincludegraphics[width=\textwidth]
        {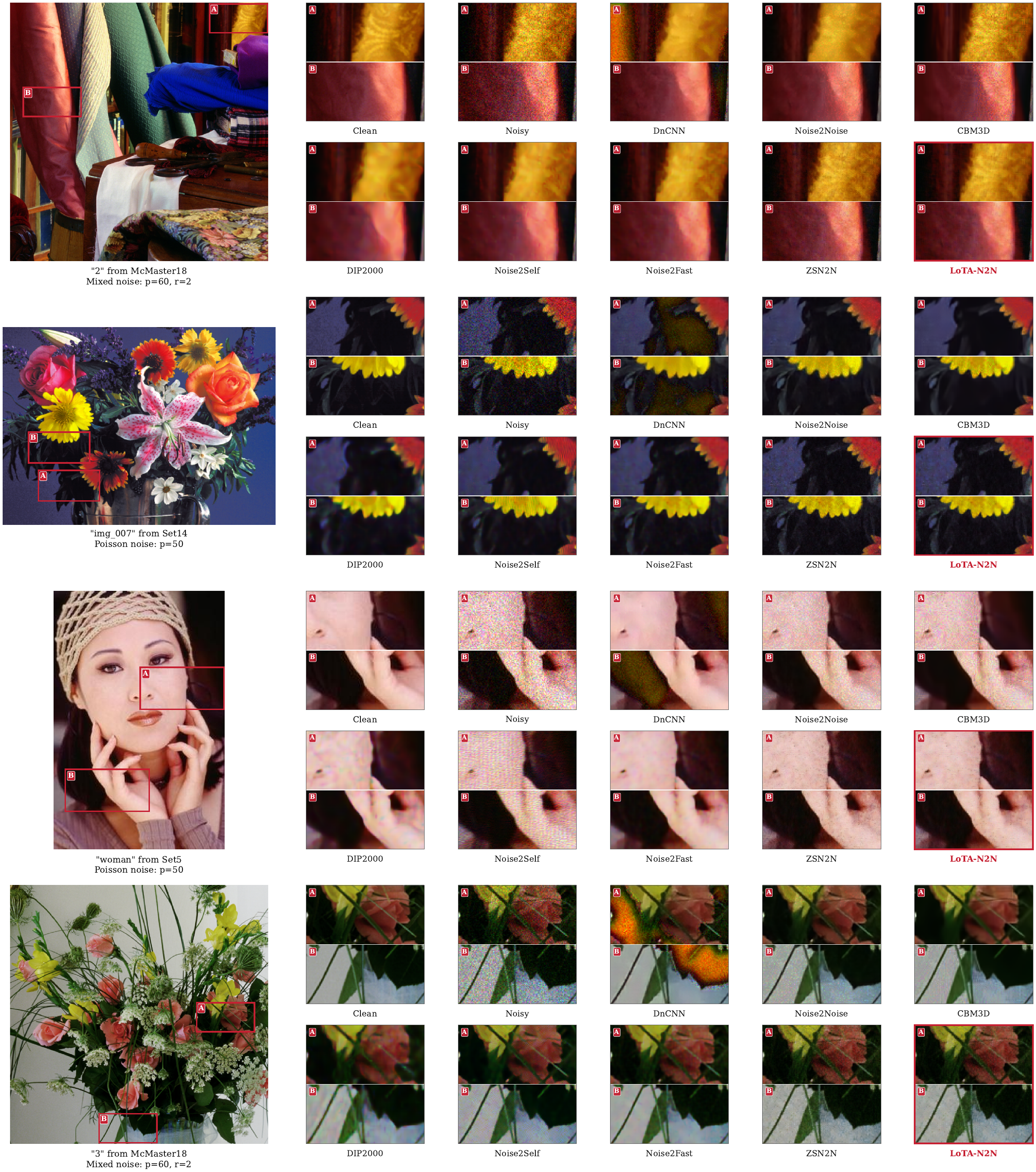}
    \caption{
    Additional qualitative comparisons on McMaster18, Set14, and Set5
    under Poisson and mixed Poisson--Gaussian noise. Enlarged regions
    include smooth fabrics and skin, dark foliage, flowers, and thin
    structures.
}
    \label{fig:extra-natural-comparison}
\end{figure*}

\begin{figure}[!t]
    \centering
    \safeincludegraphics[width=\linewidth]{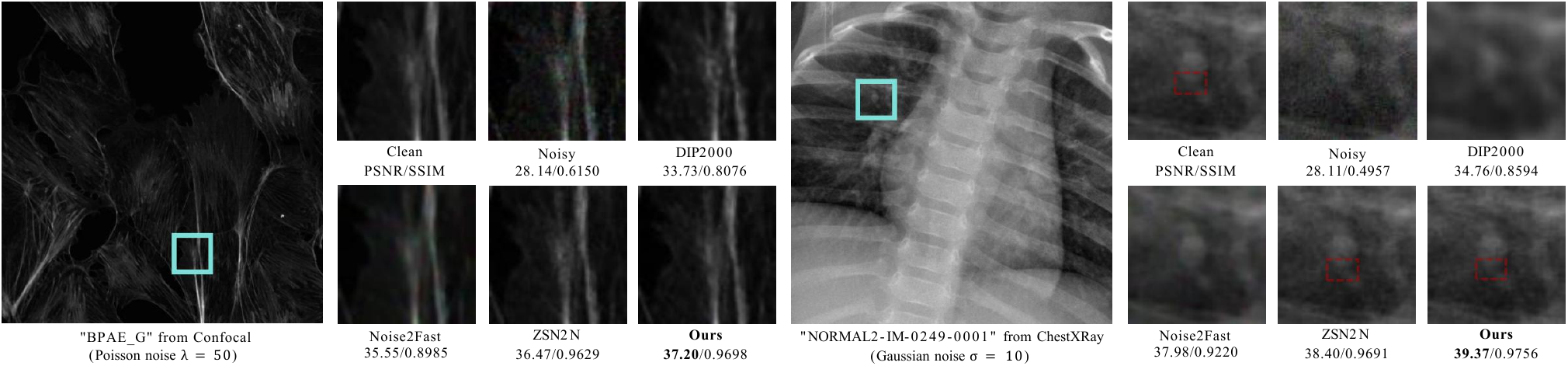}
    \caption{Qualitative comparison on confocal and X-ray images. Highlighted regions show that LoTA-N2N preserves fine biological structures and anatomical details while avoiding spurious textures.}
    \label{fig:visual-medical}
\end{figure}

Table~\ref{tab:kodak-mcmaster} reports PSNR on Kodak24, McMaster18, Set14, and
Set5 under Gaussian, Poisson, and mixed Poisson--Gaussian noise. Offline and
online methods are reported separately because they use different supervision
and adaptation regimes. Among online methods, LoTA-N2N is best in nine of the
twelve Gaussian settings and best or tied for best in all twelve Poisson
settings, giving 21 out of 24 Gaussian and Poisson settings overall. It is also
best or tied for best throughout the mixed-noise block. CBM3D remains stronger
on the three Set5 Gaussian settings, but its performance is less consistent
under Poisson and mixed corruption. Overall, LoTA-N2N provides more stable
performance across datasets, noise families, and corruption levels.

Figures~\ref{fig:visual-natural} and
\ref{fig:extra-natural-comparison} provide representative qualitative
comparisons on natural images. Across smooth regions, textured content, and
thin structures, LoTA-N2N exhibits fewer residual grains and chromatic
artifacts than DnCNN and ZSN2N while retaining the principal image boundaries.

For fairness, all rerun zero-shot methods use the same noisy inputs, image
range, clipping policy, and metric implementation. N2N is trained for 100
epochs on Kodak24 and 250 epochs on McMaster18, while DIP is run for at most
2,000 iterations. We use author-provided single-image implementations when
available and retain the published default settings unless a common
image-range conversion is required.

To examine whether the proposed objective depends on natural-image statistics,
we further evaluate LoTA-N2N on domain-specific imagery from confocal
microscopy and pediatric chest X-ray. These images differ substantially from
the natural-image benchmarks in texture, contrast, and semantic structure, but
we keep the same zero-shot adaptation protocol and do not use any external
domain-specific training data. Table~\ref{tab:confocal-xray} reports PSNR under
controlled Gaussian and Poisson corruptions, and
Figure~\ref{fig:visual-medical} shows representative reconstructions. These
experiments test cross-domain behavior under known noise rather than claiming
performance on unknown real acquisition noise.

LoTA-N2N achieves the highest PSNR in all eight confocal and X-ray settings.
Compared with ZSN2N, the gains are 0.08--0.36 dB on confocal images and
0.02--0.92 dB on X-ray images. The improvement is especially clear for
Gaussian-corrupted X-ray images, suggesting that local trace correction remains
useful when image structure and noise statistics differ from natural-image
benchmarks. These results support cross-domain generalization of the objective,
while the controlled corruption protocol avoids overclaiming real acquisition
noise robustness. The qualitative examples are consistent with the quantitative
trend and do not reveal obvious hallucinated structures.

\subsection{Statistical Reliability}
Average PSNR can conceal whether an improvement is broadly distributed or
driven by a few favorable images. We therefore compare LoTA-N2N with the
iteration-matched ZSN2N-3800 MSE control on identical noisy inputs.
Table~\ref{tab:statistics_zsn2n3800} reports the paired mean PSNR difference,
its 95\% bootstrap confidence interval, and the percentage of test images on
which LoTA-N2N performs better.

\begin{table*}[t]
    \centering
    \small
    \renewcommand{\arraystretch}{1.0}
    \setlength{\tabcolsep}{5pt}
    \begin{tabular}{l c | c c c}
        \toprule
        Dataset & Noise family
        & Matched baseline
        & $\Delta$PSNR (dB), 95\% CI
        & Win rate (\%) \tabularnewline
        \midrule
        Kodak24 & Gaussian & ZSN2N-3800$^{\ast}$ & $+0.37\;[+0.30,+0.44]$ & 91.7 \tabularnewline
        Kodak24 & Poisson & ZSN2N-3800$^{\ast}$ & $+0.39\;[+0.31,+0.47]$ & 95.8 \tabularnewline
        Kodak24 & Mixed ($r=2$) & ZSN2N-3800$^{\ast}$ & $+0.39\;[+0.31,+0.47]$ & 95.8 \tabularnewline
        McMaster18 & Gaussian & ZSN2N-3800$^{\ast}$ & $+0.50\;[+0.40,+0.61]$ & 100.0 \tabularnewline
        McMaster18 & Poisson & ZSN2N-3800$^{\ast}$ & $+0.58\;[+0.47,+0.69]$ & 100.0 \tabularnewline
        McMaster18 & Mixed ($r=2$) & ZSN2N-3800$^{\ast}$ & $+0.58\;[+0.47,+0.70]$ & 100.0 \tabularnewline
        Set14 & Gaussian & ZSN2N-3800$^{\ast}$ & $+0.19\;[-0.004,+0.37]$ & 71.4 \tabularnewline
        Set14 & Poisson & ZSN2N-3800$^{\ast}$ & $+0.31\;[+0.13,+0.47]$ & 85.7 \tabularnewline
        Set14 & Mixed ($r=2$) & ZSN2N-3800$^{\ast}$ & $+0.31\;[+0.13,+0.47]$ & 85.7 \tabularnewline
        \bottomrule
    \end{tabular}
    \caption{Paired comparison with the iteration-matched ZSN2N-3800 control. For each image, PSNR differences are averaged over the three noise levels before 10,000 image-level bootstrap resamples. ZSN2N-3800$^{\ast}$ starts from the same 3,000-step checkpoint and performs 800 additional MSE-only steps with restarted Adam and the Stage~2 learning-rate schedule. All LoTA-N2N rows use the final trace-weight floor of 0.15.}
    \label{tab:statistics_zsn2n3800}
\end{table*}

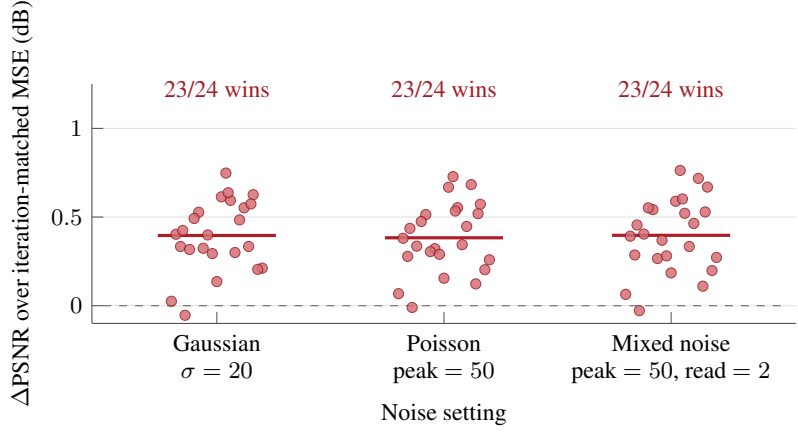
\begin{figure*}[t]
    \centering
    \begin{tikzpicture}
    \begin{axis}[
        width=0.78\textwidth,
        height=0.34\textwidth,
        ymin=-0.10,
        ymax=1.25,
        xmin=0.45,
        xmax=3.55,
        ylabel={$\Delta$PSNR over iteration-matched MSE (dB)},
        xlabel={Noise setting},
        xtick={1,2,3},
        xticklabels={
            \makecell{Gaussian\\ $\sigma=20$},
            \makecell{Poisson\\peak $=50$},
            \makecell{Mixed noise\\peak $=50$, read $=2$}
        },
        tick label style={font=\small},
        label style={font=\small},
        ymajorgrids=true,
        grid style={line width=.2pt, draw=gray!25},
        axis x line*=bottom,
        axis y line*=left,
        axis line style={draw=black!70},
        tick style={draw=black!70},
        clip=false
    ]

    \addplot[dashed, draw=black!55, line width=0.45pt]
        coordinates {(0.5,0) (3.5,0)};

    \addplot[
        only marks, mark=*, mark size=1.9pt,
        draw=lotaRed!80!black, fill=lotaRed!65, opacity=0.85
    ] coordinates {
        (0.80,0.025) (0.84,0.334) (0.88,0.317) (0.92,0.527)
        (0.96,0.399) (1.00,0.136) (1.04,0.748) (1.08,0.300)
        (1.12,0.552) (1.16,0.626) (1.20,0.212) (0.82,0.403)
        (0.86,-0.054) (0.90,0.492) (0.94,0.324) (0.98,0.294)
        (1.02,0.614) (1.06,0.594) (1.10,0.484) (1.14,0.334)
        (1.18,0.205) (0.85,0.424) (1.05,0.637) (1.15,0.574)
    };

    \addplot[
        only marks, mark=*, mark size=1.9pt,
        draw=lotaRed!80!black, fill=lotaRed!65, opacity=0.85
    ] coordinates {
        (1.80,0.068) (1.84,0.278) (1.88,0.336) (1.92,0.514)
        (1.96,0.322) (2.00,0.155) (2.04,0.728) (2.08,0.344)
        (2.12,0.683) (2.16,0.572) (2.20,0.259) (1.82,0.380)
        (1.86,-0.010) (1.90,0.475) (1.94,0.305) (1.98,0.290)
        (2.02,0.668) (2.06,0.553) (2.10,0.447) (2.14,0.123)
        (2.18,0.203) (1.85,0.436) (2.05,0.534) (2.15,0.519)
    };

    \addplot[
        only marks, mark=*, mark size=1.9pt,
        draw=lotaRed!80!black, fill=lotaRed!65, opacity=0.85
    ] coordinates {
        (2.80,0.064) (2.84,0.286) (2.88,0.404) (2.92,0.542)
        (2.96,0.369) (3.00,0.185) (3.04,0.763) (3.08,0.334)
        (3.12,0.718) (3.16,0.669) (3.20,0.272) (2.82,0.392)
        (2.86,-0.028) (2.90,0.552) (2.94,0.266) (2.98,0.281)
        (3.02,0.589) (3.06,0.521) (3.10,0.464) (3.14,0.110)
        (3.18,0.198) (2.85,0.455) (3.05,0.602) (3.15,0.529)
    };

    \addplot[mark=none, line width=1.2pt, draw=lotaRed!90!black]
        coordinates {(0.74,0.396) (1.26,0.396)};
    \addplot[mark=none, line width=1.2pt, draw=lotaRed!90!black]
        coordinates {(1.74,0.383) (2.26,0.383)};
    \addplot[mark=none, line width=1.2pt, draw=lotaRed!90!black]
        coordinates {(2.74,0.397) (3.26,0.397)};

    \node[font=\small, text=lotaRed!85!black, anchor=south]
        at (axis cs:1,1.12) {23/24 wins};
    \node[font=\small, text=lotaRed!85!black, anchor=south]
        at (axis cs:2,1.12) {23/24 wins};
    \node[font=\small, text=lotaRed!85!black, anchor=south]
        at (axis cs:3,1.12) {23/24 wins};

    \end{axis}
    \end{tikzpicture}
    \caption{Per-image Stage~2 gain of LoTA-N2N over an iteration-matched MSE control using the standard three-layer CNN on Kodak24. Both methods start from the same 3,000-step Stage~1 checkpoint and run 800 additional Stage~2 optimization steps; LoTA-N2N uses the final trace-weight floor of 0.15. Each point is one test image, and horizontal bars denote mean gains.}
    \label{fig:kodak-stage2-gain-distribution}
\end{figure*}

Figure~\ref{fig:kodak-stage2-gain-distribution} complements the three-level
bootstrap summary with per-image gains at one representative level from each
noise family.

The Kodak24 and McMaster18 confidence intervals exclude zero for all reported
noise families, with win rates of at least 91.7\% and 100\%, respectively. On
Set14, the Poisson and mixed-noise gains remain significant, whereas the
Gaussian interval includes zero at its lower endpoint. The per-image results in
Figure~\ref{fig:kodak-stage2-gain-distribution} use one representative level per
noise family and show 23/24 wins in each plotted setting; they complement, but
are not numerically identical to, the three-level averages in the table.

\subsection{Robustness under Noise-Distribution Shift}
\label{sec:stress-test}
\begin{figure*}[t]
    \centering
    \safeincludegraphics[width=0.90\textwidth]{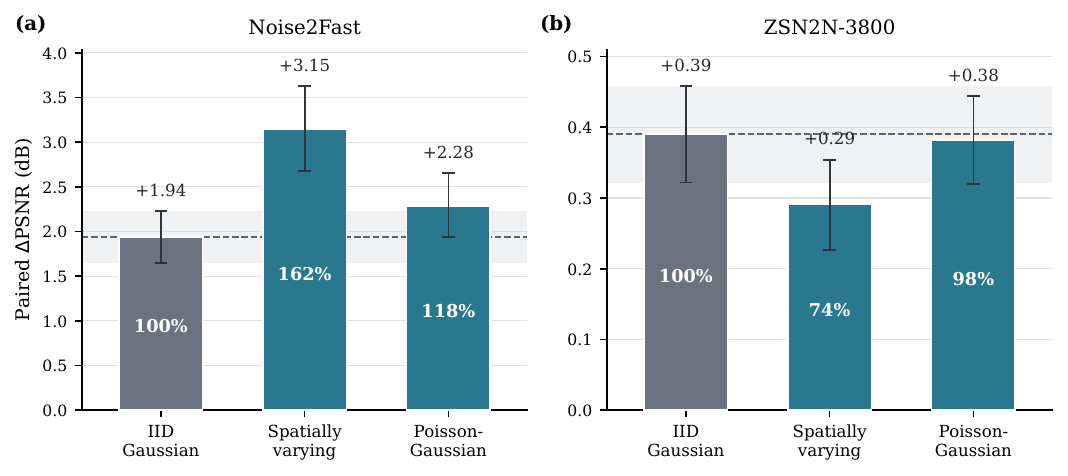}
    \caption{Retention of the floor-0.15 LoTA-N2N advantage over (a) the step-matched Noise2Fast baseline and (b) the iteration-matched ZSN2N-3800 MSE control under selected noise shifts on Kodak24. Bars show paired $\Delta$PSNR (LoTA-N2N minus the indicated baseline), and whiskers show 95\% confidence intervals from 10,000 image-level bootstrap resamples after averaging five matched noise realizations per image. Dashed lines and shaded bands mark the IID mean and its 95\% confidence interval in each panel; percentages report the advantage relative to the corresponding IID reference. Positive values favor LoTA-N2N.}
    \label{fig:noise-shift-gain-retention-joint}
\end{figure*}

The central motivation of LoTA-N2N is reduced dependence on assumptions about
target-noise independence and stationarity. We evaluate three controlled
conditions on Kodak24: IID Gaussian noise with $\sigma=25/255$, spatially
varying Gaussian noise with a horizontal standard-deviation map from $5/255$ to
$50/255$, and mixed Poisson--Gaussian noise with peak 30 followed by Gaussian
read noise with $\sigma=5/255$. No method is given the spatial noise map or the
mixed-noise parameters. Table~\ref{tab:selected-noise-evaluation} reports
absolute PSNR/SSIM, while
Figure~\ref{fig:noise-shift-gain-retention-joint} reports paired gain retention
relative to the two matched baselines.

\begin{table*}[t]
    \centering
    \small
    \renewcommand{\arraystretch}{1.0}
    \setlength{\tabcolsep}{7pt}
    \begin{tabular}{l | ccc}
        \toprule
        Method
        & IID Gaussian
        & Spatially varying
        & Poisson--Gaussian \tabularnewline
        \midrule
        Self2Self$^{\dagger}$
        & 27.61/0.7762
        & 27.21/0.7657
        & 27.13/\textbf{0.7711} \tabularnewline

        Noise2Fast
        & 27.65/0.7523
        & 25.24/0.6880
        & 26.06/0.7002 \tabularnewline

        ZSN2N
        & 29.20/0.7775
        & 28.09/0.7583
        & 27.96/0.7495 \tabularnewline

        Global TrC
        & 29.55/0.7942
        & 28.35/0.7632
        & 28.31/0.7655 \tabularnewline

        \rowcolor{oursRow}\textbf{LoTA-N2N}
        & \textbf{29.62/0.7975}
        & \textbf{28.40/0.7680}
        & \textbf{28.37}/0.7702 \tabularnewline
        \bottomrule
    \end{tabular}
    \caption{Selected non-correlated noise evaluations on Kodak24.
    Each cell reports mean PSNR (dB)/SSIM over all 24 images and
    five matched noise realizations. All optimization-based methods
    use approximately 3,800 optimization steps.
    LoTA-N2N and Global TrC use the final floor-0.15 configuration.
    $^{\dagger}$Self2Self uses a partial-convolution reimplementation
    with 100 stochastic predictions. Best results are shown in bold.}
    \label{tab:selected-noise-evaluation}
\end{table*}

Table~\ref{tab:selected-noise-evaluation} shows that LoTA-N2N obtains the best
PSNR in all three conditions and the best SSIM under IID and spatially varying
noise; Self2Self is higher by only 0.0009 SSIM under mixed noise. The paired
comparisons in Figure~\ref{fig:noise-shift-gain-retention-joint} separate
absolute task difficulty from the benefit of the proposed objective. Relative
to Noise2Fast, the LoTA-N2N advantage grows from 1.97 dB under IID noise to
3.17 dB under spatially varying noise and remains 2.31 dB under mixed noise.
Relative to ZSN2N-3800, the gains are 0.42, 0.31, and 0.41 dB, respectively.
All six image-level bootstrap confidence intervals exclude zero. Thus, the
benefit over matched baselines persists under these non-correlated shifts; this
experiment does not imply robustness to arbitrary structured or correlated
noise.

\subsection{Mechanism Validation}
\label{sec:mechanism}

The main trainable guarantee in Theorem~1 is value-level: LTrC tightens an
upper surrogate of the supervised risk. We additionally use synthetic data
with known clean images to test the first-order behavior characterized by
Proposition~1. These diagnostic quantities are never exposed to the final
training loss. At fixed checkpoints, let $\mathbf{g}_{\mathrm{sup}}$ and
$\mathbf{g}_{\mathrm{ss}}$ denote gradients of the supervised and
self-supervised objectives with respect to all trainable parameters. We report the normalized gradient error
\begin{equation}
    G_{\mathrm{rel}}
    =
    \frac{\|\mathbf{g}_{\mathrm{ss}}-\mathbf{g}_{\mathrm{sup}}\|_2}
    {\|\mathbf{g}_{\mathrm{sup}}\|_2+\delta},
\end{equation}
and their cosine similarity
\begin{equation}
    G_{\mathrm{cos}}
    =
    \frac{\langle\mathbf{g}_{\mathrm{ss}},\mathbf{g}_{\mathrm{sup}}\rangle}
    {\|\mathbf{g}_{\mathrm{ss}}\|_2\|\mathbf{g}_{\mathrm{sup}}\|_2+\delta}.
\end{equation}
The need for local trace regularization is measured with the cancellation ratio
\begin{equation}
    R_{\mathrm{cancel}}
    =
    1-
    \frac{\left|M^{-1}\sum_m\widehat{\mathcal{T}}^{(m)}\right|}
    {M^{-1}\sum_m|\widehat{\mathcal{T}}^{(m)}|+\delta}.
\end{equation}
A value near one indicates severe cancellation: the global trace appears small
even though local interactions remain large.

\begin{figure*}[t]
    \centering
    \safeincludegraphics[width=\textwidth]{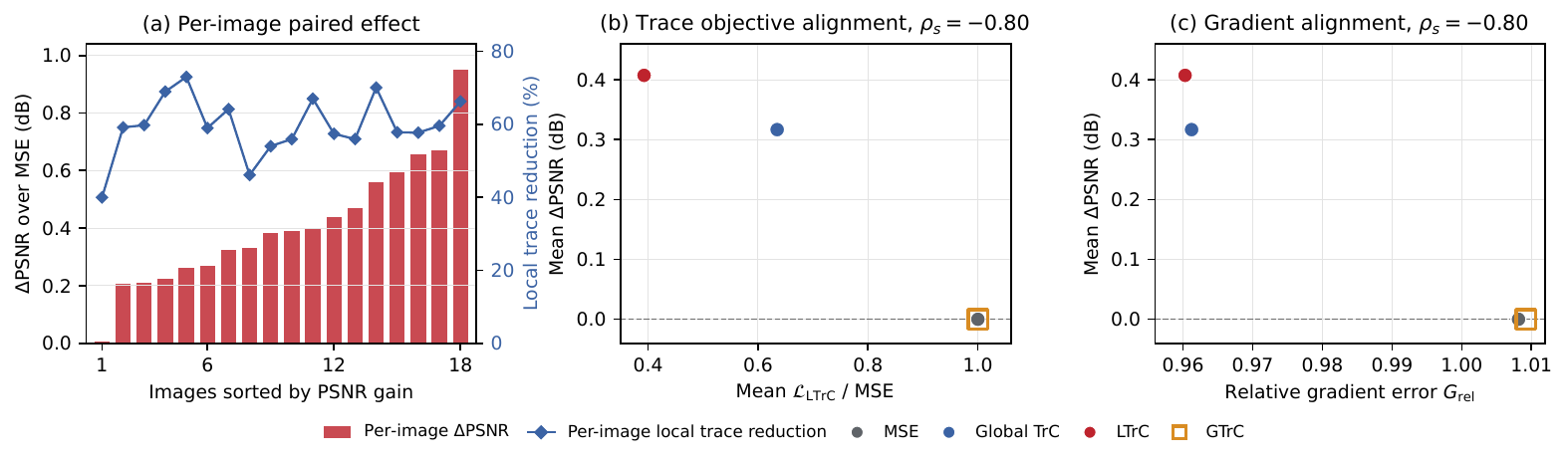}
    \caption{Mechanism evidence on McMaster18 with Gaussian noise $\sigma=10$.
    (a) Per-image PSNR gain and local-trace reduction of LTrC over the
    iteration-matched MSE control. All 18 images improve and all 18 show lower
    local trace magnitude. (b,c) Across the plotted Stage~2 objectives, lower
    mean local trace and lower relative gradient error correspond to larger
    mean PSNR gains. The hollow GTrC marker exposes its near-overlap with MSE.}
    \label{fig:mechanism}
\end{figure*}

For ablation only, we also instantiate the direct gradient-trace penalty
\begin{equation}
\begin{aligned}
    \mathcal{L}_{\mathrm{GTrC}}^{i\rightarrow j}
    &=
    \frac{1}{M}\sum_{m=1}^{M}
    \|\nabla_{\theta}
    \widehat{\mathcal{T}}_{i\rightarrow j}^{(m)}\|_2^2,\\
    \mathcal{L}_{\mathrm{GTrC}}
    &=
    \frac{1}{2}
    \sum_{(i,j)\in\{(1,2),(2,1)\}}
    \mathcal{L}_{\mathrm{GTrC}}^{i\rightarrow j}.
\end{aligned}
\end{equation}
It requires double backpropagation and is not part of LoTA-N2N. Table~\ref{tab:mechanism-ablation}
compares this diagnostic penalty with the final LTrC objective, and
Figure~\ref{fig:mechanism} visualizes the corresponding per-image and
mechanism-level relationships.

\begin{table*}[t]
    \centering
    \small
    \renewcommand{\arraystretch}{1.0}
    \setlength{\tabcolsep}{3.5pt}
    \begin{tabular}{l | c c c c c}
        \toprule
        Stage~2 objective
        & PSNR/SSIM
        & $|\widehat{\mathcal{T}}|$
        & $\mathcal{L}_{\mathrm{LTrC}}$
        & $G_{\mathrm{rel}}\downarrow$
        & $G_{\mathrm{cos}}\uparrow$ \tabularnewline
        \midrule
        No Stage~2
        & 34.14/0.8900 & -- & -- & -- & -- \tabularnewline
        MSE, iteration-matched
        & 34.17/0.8912 & $4.11{\times}10^{-5}$ & $5.14{\times}10^{-5}$
        & 1.008 & 0.058 \tabularnewline
        MSE $+$ Global TrC
        & 34.49/0.8985 & $5.53{\times}10^{-6}$ & $3.26{\times}10^{-5}$
        & 0.961 & 0.251 \tabularnewline
        \rowcolor{oursRow}\textbf{MSE $+$ LTrC}
        & 34.58/0.9027 & $1.37{\times}10^{-5}$ & $2.02{\times}10^{-5}$
        & 0.960 & 0.164 \tabularnewline
        MSE $+$ GTrC
        & 34.17/0.8912 & $4.11{\times}10^{-5}$ & $5.14{\times}10^{-5}$
        & 1.009 & 0.053 \tabularnewline
        MSE $+$ LTrC $+$ GTrC
        & 34.58/0.9026 & $1.37{\times}10^{-5}$ & $2.02{\times}10^{-5}$
        & 0.959 & 0.171 \tabularnewline
        Oracle trace diagnostic$^{\dagger}$
        & --/-- & $|\mathcal{T}|=1.00{\times}10^{-4}$
        & $E_{\mathcal{T}}=1.169$ & -- & -- \tabularnewline
        \bottomrule
    \end{tabular}
    \caption{Objective ablation and mechanism diagnostics on McMaster18 with
    Gaussian noise $\sigma=10$. Entries are means over all 18 images, and all
    Stage~2 trace regularizers use a floor of 0.15. The MSE control uses
    exactly the same Stage~2 iterations as the trace-regularized variants.
    Rows containing GTrC are diagnostic ablations and are not part of the final
    method. $^{\dagger}$The oracle row reports the mean absolute clean-image
    trace and mean relative trace-estimation error at the final LTrC checkpoint;
    it is a diagnostic rather than a trainable zero-shot method.}
    \label{tab:mechanism-ablation}
\end{table*}

\begin{table*}[t]
    \centering
    \small
    \renewcommand{\arraystretch}{1.0}
    \setlength{\tabcolsep}{5pt}
    \resizebox{\textwidth}{!}{%
    \begin{tabular}{c | cc | cccc}
        \toprule
        Setting
        & Bidirectional
        & Residual
        & $\sigma=10$
        & $\sigma=15$
        & $\sigma=20$
        & $\sigma=25$ \tabularnewline
        \midrule
        $A_1$ & \ding{56} & \ding{56} & 33.7612/0.8826 & 31.5647/0.8262 & 29.9177/0.7735 & 28.5585/0.7199 \\
        $A_2$ & \ding{52} & \ding{56} & 34.0974/0.8862 & 32.0574/0.8429 & 30.5368/0.8039 & 29.2653/0.7660 \\
        $A_3$ & \ding{56} & \ding{52} & 34.1347/0.8944 & 31.8502/0.8380 & 30.1274/0.7817 & 28.7153/0.7282 \\
        \rowcolor{oursRow} $\mathbf{A}_4$ & \ding{52} & \ding{52} & \textbf{34.5810/0.9026} & \textbf{32.4209/0.8585} & \textbf{30.7546/0.8156} & \textbf{29.3918/0.7748} \\
        \bottomrule
    \end{tabular}%
    }
    \caption{Design ablation on McMaster18 under the final LTrC-only method.
    Entries report mean PSNR/SSIM over all 18 images using one fixed noise
    realization per image and optimization seed 2027. Every setting trains its
    own Stage~1 model and then receives the same 800-step floor-0.15 Stage~2 update with
    $\mathcal{L}_{\mathrm{MSE}}+\lambda_{\mathrm{v}}
    \mathcal{L}_{\mathrm{LTrC}}$.}
    \label{tab:design-ablation}
\end{table*}

Table~\ref{tab:mechanism-ablation} provides the controlled objective ablation.
The iteration-matched MSE row separates the effect of LTrC from additional
optimization, while Global TrC and LTrC isolate global value suppression from
local cancellation control. The GTrC rows test whether directly penalizing the
trace gradient adds value beyond the final objective. It does not: GTrC alone
is indistinguishable from MSE, and adding it to LTrC changes PSNR by only
$+0.0001$ dB. In contrast, LTrC substantially reduces local trace magnitude and
improves PSNR, supporting Proposition~3 and the upper-bound construction in
Theorem~1.

Figure~\ref{fig:mechanism} further tests whether the reconstruction gain follows
the proposed mechanism rather than merely adding another regularizer. Panel~(a)
compares LTrC with the iteration-matched MSE update from the same Stage~1
checkpoint and optimization budget. Every image improves in PSNR while its
local trace magnitude decreases, indicating that the effect is not driven by a
few favorable outliers. Panels~(b,c) compare Stage~2 objectives and show that smaller local trace and
smaller diagnostic gradient mismatch generally accompany larger mean PSNR
gains. The key causal evidence is the matched LTrC comparison in panel~(a);
the near-overlap of GTrC with MSE confirms that an explicit gradient penalty is
unnecessary. These results are consistent with LTrC tightening the supervised
risk surrogate, while improved first-order alignment is treated as an observed
diagnostic rather than a universal guarantee.

\begin{figure*}[t]
    \centering
    \safeincludegraphics[width=0.92\textwidth]{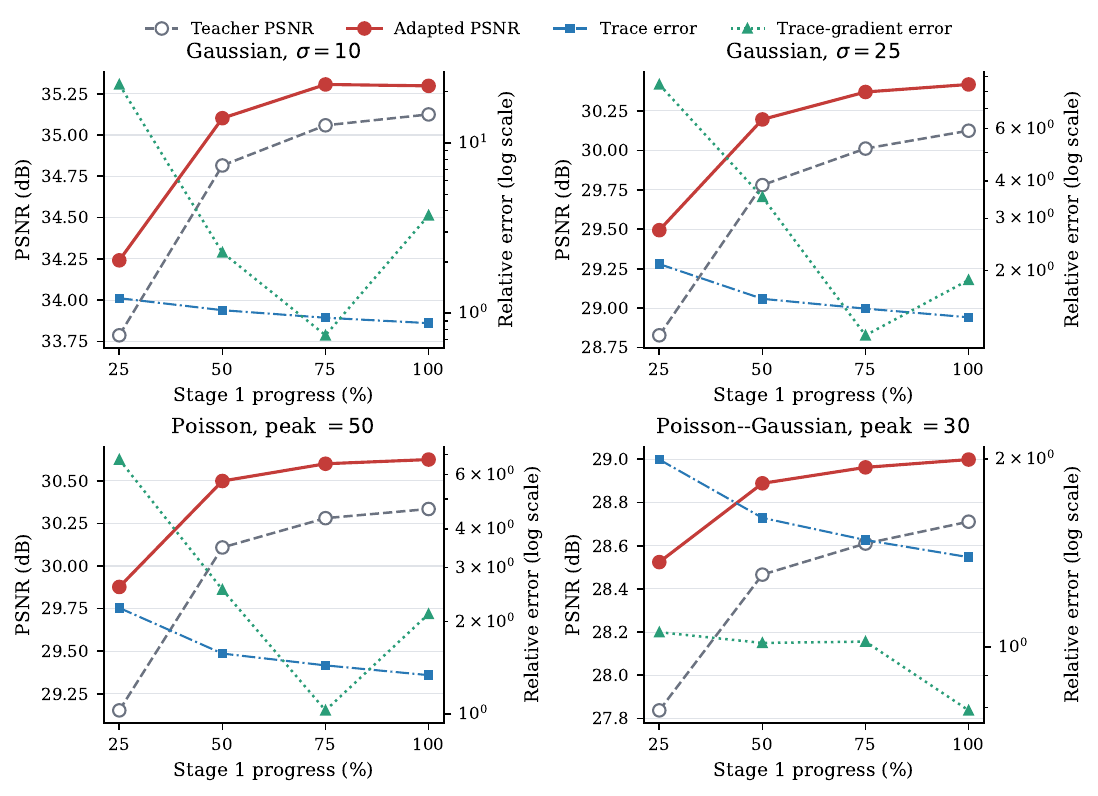}
    \caption{Effect of Stage~1 teacher quality on trace-estimation reliability.
    We vary the frozen Stage~1 checkpoint used to construct detached clean
    sub-image estimates on Set12 validation images. Across Gaussian, Poisson,
    and mixed noise settings, stronger teachers generally yield higher adapted
    PSNR and lower relative trace-estimation error, consistent with
    Proposition~2 and the estimation-error term in Theorem~1. Every adapted
    model uses the final 800-step MSE$+$LTrC Stage~2 objective with floor 0.15;
    trace-gradient error is computed only after training as a diagnostic and is
    never optimized.}
    \label{fig:teacher-quality}
\end{figure*}

Figure~\ref{fig:teacher-quality} next tests the estimation-error term in
Theorem~1 by varying the quality of the frozen Stage~1 teacher.

We also examine the reliability assumption behind the estimated trace by
varying the quality of the frozen Stage~1 teacher. Figure~\ref{fig:teacher-quality}
uses intermediate Stage~1 checkpoints to construct the detached clean estimates
while keeping the Stage~2 protocol fixed. As the teacher improves, the final
adapted PSNR generally rises, and the relative error of the estimated trace
decreases across the reported noise settings. Across the four corruption
settings, the mean per-image Spearman correlation between teacher PSNR and
trace-estimation error ranges from $-0.84$ to $-1.00$, while its correlation
with adapted PSNR ranges from $0.76$ to $1.00$. This supports the error bound in
Proposition~2: LoTA-N2N does not require exact clean targets, but its correction
is most reliable when the Stage~1 denoiser provides a sufficiently accurate
proxy. The non-monotonic trace-gradient curves in a few settings also reinforce
the theoretical distinction made after Proposition~1: better value estimation
does not always imply monotone first-order behavior. This is why the final
method optimizes LTrC alone and reports gradient quantities only as diagnostics.

\subsection{Design Ablation}
\begin{figure*}[t]
    \centering
    \safeincludegraphics[width=0.92\textwidth]{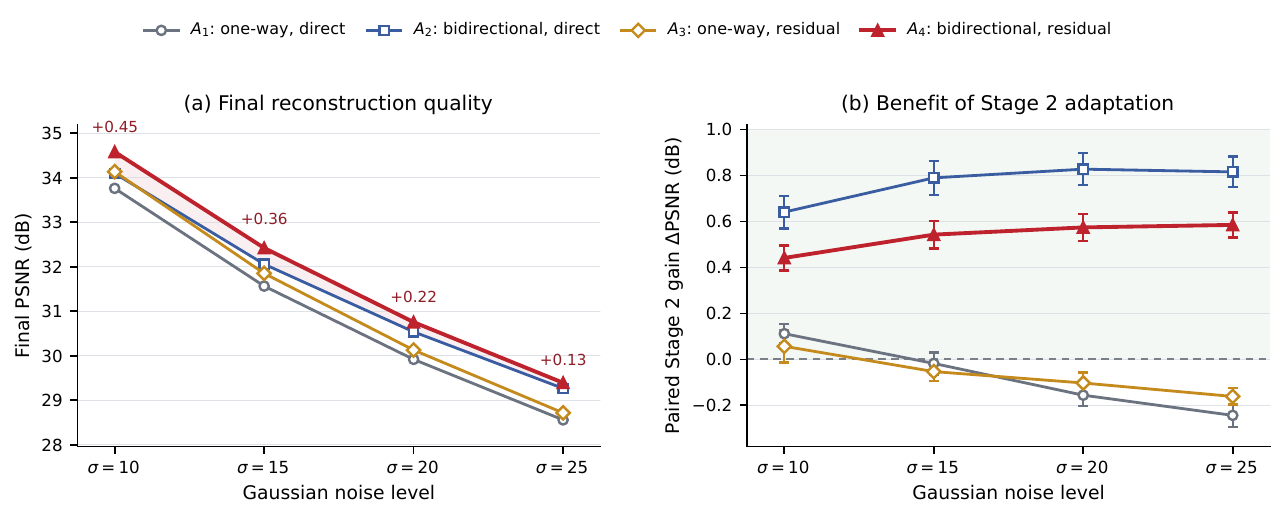}
    \caption{Design ablation under the final floor-0.15 LTrC-only objective on
    McMaster18. Panel~(a) compares final PSNR across Gaussian noise levels, with
    annotations reporting the margin of $A_4$ over the strongest alternative.
    Panel~(b) reports the paired Stage~2 PSNR gain with standard-error bars.
    Bidirectional training produces positive Stage~2 gains at every noise level,
    and $A_4$, which also uses residual prediction, obtains the highest final
    PSNR in all four settings.}
    \label{fig:design-ablation}
\end{figure*}

After validating the final LTrC objective, we isolate two implementation choices
that are orthogonal to the trace formulation: bidirectional sub-image training
and residual prediction. Each setting in Table~\ref{tab:design-ablation}
trains its corresponding Stage~1 model and then uses the same final 800-step
MSE$+$LTrC Stage~2 protocol. Figure~\ref{fig:design-ablation} visualizes the
corresponding Stage~1 initialization and Stage~2 improvement.

Table~\ref{tab:design-ablation} and
Figure~\ref{fig:design-ablation} show complementary effects from
bidirectionality and residual prediction. The bidirectional variants $A_2$ and
$A_4$ improve at every noise level, whereas the one-way variants can degrade
under stronger corruption. $A_4$ achieves the highest final PSNR at each noise
level. Because every row
uses the same final LTrC-only objective and optimization budget, the comparison
isolates the two architectural choices rather than changes in the loss.

\subsection{Sensitivity and Computational Cost}
\begin{figure*}[t]
    \centering
    \safeincludegraphics[width=\textwidth]{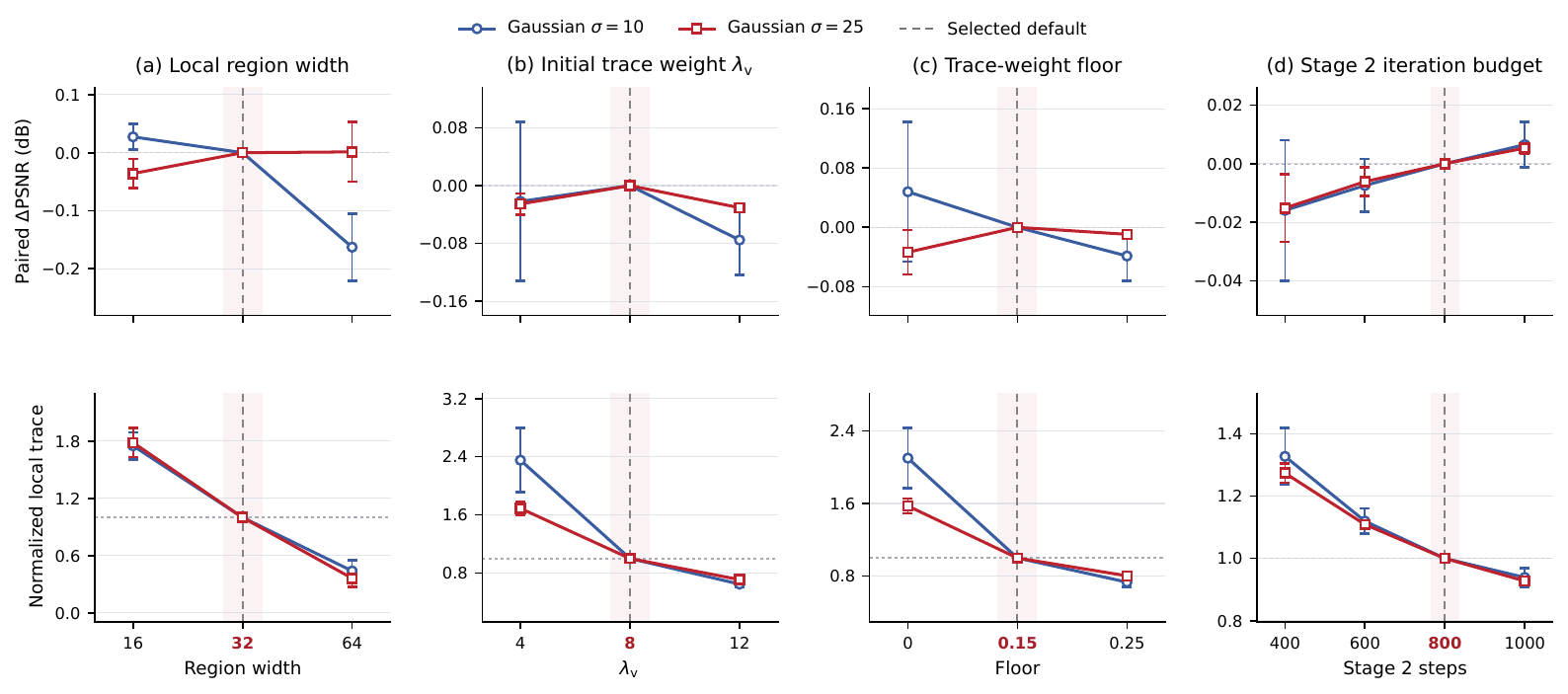}
    \caption{One-factor sensitivity analysis of the final LTrC-only
    configuration on five Set12 validation images under Gaussian noise with
    $\sigma=10$ and $\sigma=25$. Columns vary local region width, the initial
    LTrC weight $\lambda_{\mathrm{v}}$, the trace-weight floor, and the Stage~2
    iteration budget. The top row reports paired PSNR changes relative to the
    selected default, and the bottom row reports local-trace magnitude normalized
    by the same per-image default. Curves show five-image means with standard-error
    bars; vertical dashed lines and highlighted tick labels mark the final settings.}
    \label{fig:sensitivity}
\end{figure*}

We use five Set12 validation images, disjoint from every reported benchmark, to
study one-factor sensitivity of the final LTrC-only configuration. The
sweep varies the local region width, initial trace weight
$\lambda_{\mathrm{v}}$, trace-weight floor, and Stage~2 iteration budget while
holding the remaining settings fixed. Because the absolute PSNR gap between
$\sigma=10$ and $\sigma=25$ would obscure the much smaller hyperparameter
effects, Figure~\ref{fig:sensitivity} reports paired PSNR changes relative to
the selected default. A normalized local-trace value below one indicates
stronger trace suppression than that default. The corresponding end-to-end
cost comparison is reported in Table~\ref{tab:runtime-scaling-3800}.

Figure~\ref{fig:sensitivity} shows that the final configuration is stable but
not insensitive to arbitrarily aggressive trace suppression. A region width of
32 provides the most balanced behavior: increasing it to 64 reduces normalized
local trace to 0.44 and 0.36 for $\sigma=10$ and $\sigma=25$, respectively,
but decreases PSNR by 0.16 dB at $\sigma=10$. The selected
$\lambda_{\mathrm{v}}=8$ is the best of the three tested weights at both noise
levels; increasing it to 12 further suppresses trace but reduces PSNR by
0.08/0.03 dB. The floor sweep exposes the same trade-off. Removing the floor
improves $\sigma=10$ by 0.05 dB but degrades $\sigma=25$ by 0.03 dB and raises
the normalized trace to 2.10/1.57, whereas a floor of 0.25 provides stronger
suppression without improving PSNR. Thus, 0.15 is a noise-level-robust
compromise rather than a value selected for one favorable condition.

Performance also saturates with the Stage~2 budget. Relative to 800 steps,
400 and 600 steps are lower by at most 0.016 and 0.008 dB, while extending the
budget to 1,000 steps gains only 0.006 dB on average and increases Stage~2 time
by approximately 25\%. We therefore use region width 32,
$\lambda_{\mathrm{v}}=8$, floor 0.15, and 800 Stage~2 steps in the final
configuration. These settings avoid the PSNR loss caused by over-suppressing
the estimated interaction while retaining most of the attainable trace
reduction.

\begin{table*}[t]
    \centering
    \footnotesize
    \renewcommand{\arraystretch}{1.05}
    \setlength{\tabcolsep}{3.2pt}
    \resizebox{\textwidth}{!}{%
    \begin{tabular}{l c c | cc | cc | cc}
        \toprule
        \multirow{2}{*}{Method}
        & \multirow{2}{*}{Steps/image}
        & \multirow{2}{*}{Params (M)}
        & \multicolumn{2}{c|}{$125\times125$}
        & \multicolumn{2}{c|}{$250\times250$}
        & \multicolumn{2}{c}{$500\times500$} \\
        \cmidrule(lr){4-5}
        \cmidrule(lr){6-7}
        \cmidrule(lr){8-9}
        & & & Time (s) & Mem. (GB) & Time (s) & Mem. (GB) & Time (s) & Mem. (GB) \\
        \midrule
        Noise2Fast & $\approx 3801$ & 0.259 & 22.11 & 0.03 & 25.19 & 0.09 & 33.61 & 0.34 \\
        ZSN2N & 3800 & 0.022 & 16.51 & 0.02 & 18.68 & 0.07 & 20.24 & 0.28 \\
        Self2Self$^{\dagger}$ & 3800 & 0.991 & 45.82 & 6.91 & 44.41 & 15.16 & 69.23 & 10.66 \\
        Global TrC & 3800 & 0.022 & 18.19 & 0.02 & 18.12 & 0.07 & 19.85 & 0.28 \\
        \rowcolor{oursRow}\textbf{LoTA-N2N} & 3800 & 0.022 & 18.81 & 0.02 & 18.63 & 0.07 & 20.37 & 0.28 \\
        \bottomrule
    \end{tabular}%
    }
    \caption{Resolution-scaling runtime and memory comparison on an NVIDIA H100
    80GB HBM3 using random RGB inputs. Each entry reports the average over three
    random inputs. The LoTA-N2N timing uses the final floor-0.15 Stage~2
    configuration. Noise2Fast is trained for 1267 steps per channel,
    corresponding to approximately 3801 total optimization steps per RGB
    image. PyTorch methods report peak
    \texttt{torch.cuda.max\_memory\_allocated}; Self2Self memory is measured by
    peak \texttt{nvidia-smi} sampling because TensorFlow compat-v1 does not
    expose the same allocator statistic. $^{\dagger}$Self2Self is measured with
    a TensorFlow compat-v1 reimplementation of the partial-convolution
    Self2Self runtime and includes 100 final stochastic predictions; its
    non-trainable output averaging buffer is excluded from the parameter count.}
    \label{tab:runtime-scaling-3800}
\end{table*}

Taken together, Figure~\ref{fig:sensitivity} and
Table~\ref{tab:runtime-scaling-3800} show that LoTA-N2N does not rely on a
single sharply tuned hyperparameter value. The selected defaults lie in the
stable region of the performance--trace trade-off. LTrC also adds no model
parameters and only a small runtime increment over ZSN2N at $500\times500$
(20.37 versus 20.24 seconds in this benchmark). The final method therefore
retains the computational profile of the lightweight baseline while avoiding
the double-backpropagation cost of GTrC.

\subsection{Transfer Across Denoising Backbones}
To test whether the improvement is architecture-specific, we repeat the iteration-matched MSE and LoTA-N2N objectives with three backbones: the default three-layer CNN, an eight-layer residual CNN, and a compact encoder--decoder. The same loss weights and Stage~2 iteration budget are used for all
backbones. Table~\ref{tab:backbone-transfer} reports the within-backbone MSE
and LoTA-N2N comparison together with parameter counts and normalized runtime.

\begin{table*}[t]
    \centering
    \footnotesize
    \renewcommand{\arraystretch}{1.0}
    \setlength{\tabcolsep}{3pt}
    \resizebox{\textwidth}{!}{%
    \begin{tabular}{l | c | cc | cc | cc | cc | c}
        \toprule
        Backbone & Params (M)
        & \multicolumn{2}{c|}{Gaussian}
        & \multicolumn{2}{c|}{Poisson}
        & \multicolumn{2}{c|}{Mixed Noise}
        & \multicolumn{2}{c|}{Correlated Gaussian}
        & Time overhead \tabularnewline
        \cmidrule(lr){3-4}
        \cmidrule(lr){5-6}
        \cmidrule(lr){7-8}
        \cmidrule(lr){9-10}
        & & MSE & LoTA-N2N & MSE & LoTA-N2N & MSE & LoTA-N2N
        & MSE & LoTA-N2N & vs. standard \tabularnewline
        \midrule
        Three-layer CNN (standard)
        & 0.022
        & 30.19 & \textbf{30.77}
        & 30.28 & \textbf{30.84}
        & 28.11 & \textbf{28.71}
        & \textbf{23.08} & 22.98
        & 1.00$\times$ \\

        Eight-layer residual CNN
        & 0.126
        & 30.52 & \textbf{30.60}
        & 30.32 & \textbf{30.52}
        & 28.52 & \textbf{28.57}
        & 23.36 & \textbf{23.38}
        & 2.88$\times$ (+188.4\%) \\

        Compact encoder--decoder
        & 0.053
        & 30.85 & \textbf{31.00}
        & 30.98 & \textbf{31.08}
        & 28.97 & \textbf{29.00}
        & \textbf{23.12} & 23.00
        & 1.57$\times$ (+57.5\%) \\
        \bottomrule
    \end{tabular}%
    }
    \caption{Transfer of the proposed objective across denoising backbones on McMaster18. Entries report PSNR (dB) for Gaussian noise with $\sigma=20$, Poisson noise with peak value 50, mixed Poisson--Gaussian noise with peak value 50 plus Gaussian noise $\sigma=20$, and correlated Gaussian noise with $\sigma=20$ using a $3\times3$ kernel with kernel standard deviation 0.8. All LoTA-N2N entries use the final floor-0.15 configuration. Runtime is normalized by the standard three-layer CNN configuration.}
    \label{tab:backbone-transfer}
\end{table*}

Table~\ref{tab:backbone-transfer} shows positive mean LoTA-N2N gains over the
iteration-matched MSE objective for every backbone under IID Gaussian, Poisson,
and mixed noise. The gain is largest for the standard three-layer CNN
(0.56--0.60 dB), while the deeper residual CNN improves by 0.05--0.20 dB and
the compact encoder--decoder by 0.03--0.15 dB. Under correlated Gaussian noise,
however, the standard and compact backbones decrease by 0.10 and 0.12 dB,
respectively, while the residual backbone improves by only 0.02 dB. Thus, the
objective transfers across architectures for the supported noise families, but
spatial correlation remains a consistent boundary. The within-backbone
comparison is the relevant quantity; absolute rankings across architectures
also reflect differences in capacity and runtime.

\section{Discussion and Limitations}
\label{sec:limitations}
The analysis and experiments identify several limitations. First,
Proposition~2 shows that an inaccurate Stage~1 teacher yields an inaccurate
trace estimate; highly textured images or extreme corruption may therefore
limit Stage~2. Second, the selected robustness study covers nonstationary and
mixed noise but does not claim general robustness to spatial correlation. In
the floor-0.15 correlated-Gaussian diagnostic in
Table~\ref{tab:backbone-transfer}, LoTA-N2N is 0.10--0.12 dB below the
iteration-matched MSE control for the standard and compact backbones and gains
only 0.02 dB for the deeper residual backbone. Correlated noise is therefore a
documented failure boundary, not a supported robustness claim.
Third, Proposition~1 shows that small trace values do not universally imply
small trace gradients without additional regularity. The final method therefore
claims a value-level supervised-risk bound, while gradient alignment is reported
only as a diagnostic. Direct GTrC regularization requires double backpropagation
and provides no PSNR benefit in our validation and ablation results, so it is
excluded from LoTA-N2N. Finally, per-image adaptation remains slower than a
pretrained feed-forward denoiser and is most appropriate when external training
data or a reliable noise model is unavailable. The paired confidence intervals and per-image plots reduce, but do
not eliminate, uncertainty about behavior on other camera pipelines.

\section{Conclusion}
\label{sec6}

We presented a trace-based analysis of MSE self-supervision that distinguishes
objective-value mismatch from first-order optimization discrepancy and unifies
several existing objectives as indirect mechanisms for suppressing the same
interaction. This analysis motivated LoTA-N2N, which combines detached
clean-target estimation with patch-wise local trace adaptation. The final method
uses only MSE and LTrC: local trace magnitudes prevent spatial cancellation and
upper-bound the inaccessible supervised risk up to teacher-estimation error.
Gradient quantities are retained only as diagnostics, and the explicit GTrC
ablation confirms that direct gradient penalization is unnecessary for
reconstruction. Among online methods, LoTA-N2N is best or tied for best in
21 of the 24 Gaussian and Poisson settings and remains competitive throughout
the mixed-noise block. Although CBM3D is stronger in a few Gaussian cases,
LoTA-N2N is more stable across datasets, noise families, and corruption levels.
It also improves over iteration-matched MSE controls across the reported
standard, spatially varying, and mixed-noise settings, while the correlated-noise
result exposes a clear boundary. Beyond the specific
denoising model, the trace perspective provides a principled framework for
analyzing surrogate objectives under weaker, but not absent, assumptions.

\bibliographystyle{iclr2026_conference}
\bibliography{ref}
\end{document}